\documentclass[10pt,twocolumn,letterpaper]{article}

\usepackage{wacv}
\usepackage{multirow}
\usepackage{array}
\usepackage{booktabs}
\usepackage{colortbl}
\usepackage{pifont}
\usepackage{adjustbox}

\usepackage{algorithm}
\usepackage{algpseudocode}
\usepackage{listings}

\usepackage{xcolor}

\definecolor{wacvblue}{rgb}{0.21,0.49,0.74}
\usepackage[pagebackref,breaklinks,colorlinks,allcolors=wacvblue]{hyperref}
\usepackage{microtype}
\def\wacvPaperID{316}
\def\confName{WACV}
\def\confYear{2027}

\title{Attention Maps Know When to Think: \\ Training-Free Difficulty-Aware Chain-of-Thought for Reasoning Segmentation}

\author{Chanseul Cho\textsuperscript{\ddag} \quad Jaesung Jun\textsuperscript{\ddag} \quad Dongheon Lee \quad Youngmin Ro\textsuperscript{*} \\
Machine Intelligence Laboratory, University of Seoul \\
Data Science Interdisciplinary Major \\
\url{https://github.com/chanseul01/DA-CoT}
}

\begin{document}
\AddToHookNext{shipout/background}{%
  \put(0,-\paperheight){\makebox[\paperwidth]{\raisebox{0.3in}{\parbox{0.8\paperwidth}{\centering\scriptsize
    \copyright~2027 IEEE. Personal use of this material is permitted. Permission from IEEE must be obtained for all other uses, in any current or future media, including reprinting/republishing this material for advertising or promotional purposes, creating new collective works, for resale or redistribution to servers or lists, or reuse of any copyrighted component of this work in other works.}}}}}
\maketitle
\renewcommand{\thefootnote}{\fnsymbol{footnote}}
\footnotetext[1]{Corresponding author.}
\footnotetext[3]{Equal contribution.}
\renewcommand{\thefootnote}{\arabic{footnote}}
\begin{abstract}
Chain-of-Thought~(CoT) prompting has become a standard reasoning strategy in Multi-modal Large Language Models~(MLLMs), showing strong performance in complex vision-language tasks such as reasoning segmentation.
However, our analysis reveals that indiscriminate CoT usage can lead to degraded performance by inducing unnecessary reasoning, a phenomenon we term overthinking.
In reasoning segmentation, overthinking often leads to misalignment between the query’s intent and visual grounding.
To address this, we propose Difficulty-Aware Chain-of-Thought~(DA-CoT), a training-free adaptive framework that selectively triggers CoT based on the input’s intrinsic difficulty.
We find that the entropy of the model’s attention maps reliably reflects task difficulty: low entropy indicates confident, localized grounding, while high entropy suggests uncertainty and potential benefit from CoT.
DA-CoT uses this entropy signal to determine whether to apply CoT prompting.
Additionally, we introduce a Bidirectional Rethinking mechanism that enables the model to automatically flip its reasoning decision when the initial output fails, enhancing robustness by recovering from failures induced by overthinking and underthinking.
DA-CoT demonstrates its effectiveness by achieving state-of-the-art performance on the Reasoning Segmentation benchmarks while significantly improving inference efficiency.

\end{abstract}

\vspace{-6mm}

\section{Introduction}
\vspace{-2mm}
\begin{figure}[t]
\centering
\includegraphics[width=1.0\linewidth]{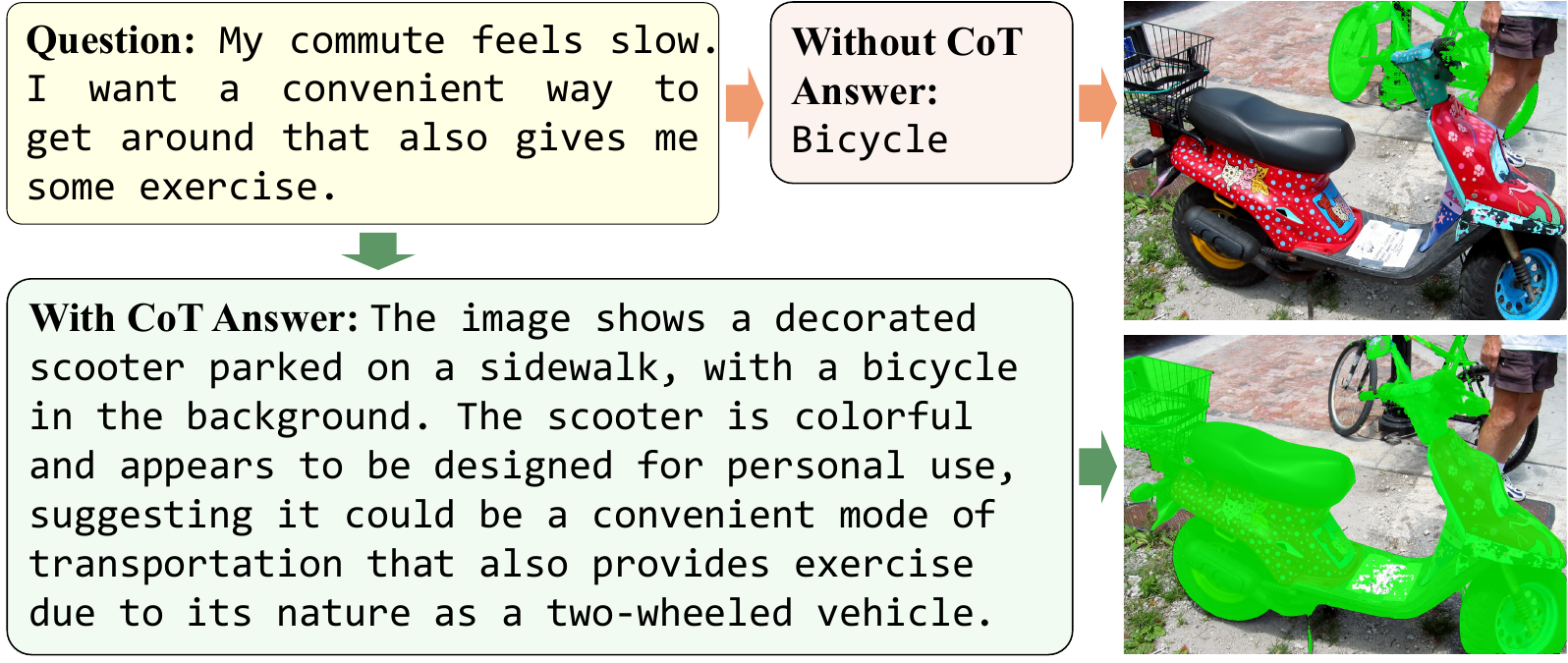}
\vspace{-6mm}
\caption{Comparison of pretrained MLLMs applied to reasoning segmentation in a training-free manner, with and without Chain-of-Thought (CoT) prompting. The CoT-prompted model fails due to overthinking caused by excessive spatial reasoning, while the model without CoT correctly identifies the bicycle.}
\label{fig:teaser}
\vspace{-8mm}
\end{figure}

\begin{figure*}[t]
\vspace{-6mm}
\centering
\includegraphics[width=\linewidth]
{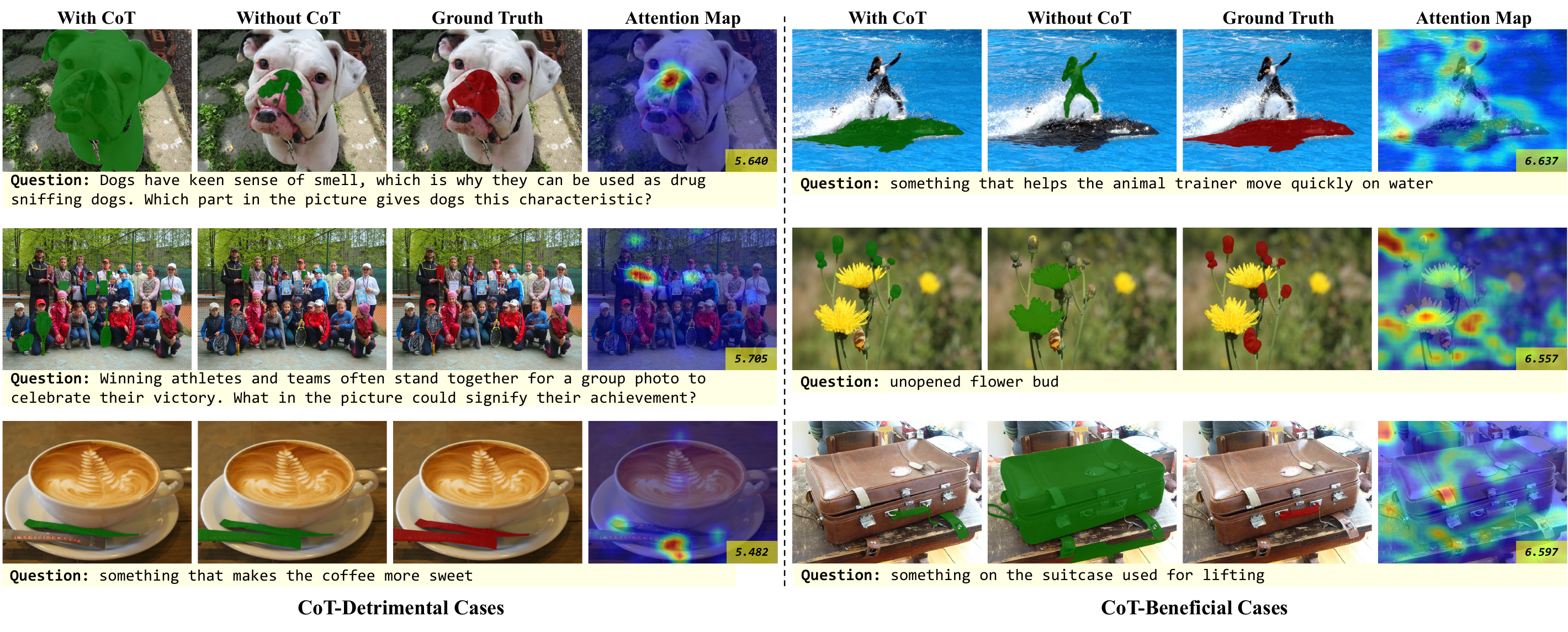}
\vspace{-8mm}
\caption{Qualitative examples illustrating the contrasting effects of CoT prompting in reasoning segmentation.
\textbf{Left}: CoT-detrimental cases, where CoT leads to misgrounding by overthinking spatial reasoning; without CoT, the model correctly identifies the target with concentrated attention and low entropy.
\textbf{Right}: CoT-beneficial cases, where CoT enhances performance in ambiguous scenarios by guiding reasoning towards the correct region, as reflected in high-entropy attention maps.
The visualized attention entropy (bottom right of each map) provides an intrinsic signal that distinguishes when CoT is helpful or harmful.}
\label{fig:attention_map}
\vspace{-6mm}
\end{figure*}

Multi-modal Large Language Models (MLLMs) have demonstrated strong performance on a wide range of vision-language tasks, largely attributed to their advanced reasoning capabilities.
Among various prompting strategies, Chain-of-Thought (CoT) prompting has emerged as a powerful technique for eliciting step-by-step reasoning, enabling models to break down complex queries into intermediate inferences~\cite{cot}.
As a result, CoT has become a default reasoning framework across vision-language tasks, including complex visual question answering, referring expression comprehension, and reasoning segmentation—a task requiring both spatial understanding and logical deduction ~\cite{segzero, learntoexplain, vctp, vqa_cot, vqa_cot2}.
Reasoning segmentation has recently emerged as a key benchmark for evaluating how well models align visual perception with high-level reasoning, and recent approaches have incorporated CoT prompting to achieve highly competitive results~\cite{rsvp, samr1, visionreasoner, sam-veteran}.

However, in line with recent evidence that additional reasoning does not universally benefit language and multimodal models~\cite{adaptive_thought, cotornot, vapo}, our empirical findings challenge this prevailing trend in reasoning segmentation: we observe that indiscriminate use of CoT prompting can lead to degraded performance due to \textit{overthinking}, where the model engages in unnecessary or misleading reasoning steps that distort the grounding process and lead to incorrect predictions.
As shown in Figure~\ref{fig:teaser}, the model correctly identifies the target region without CoT, but CoT prompting flips a correct localization into a misgrounded prediction by inducing unnecessary spatial reasoning that dilutes the model’s focus on the query’s intent.
These findings reveal that, in reasoning segmentation, unnecessary CoT reasoning can compromise the alignment between linguistic intent and visual evidence, often leading to performance degradation. We provide a quantitative breakdown of these failure modes in the supplementary material (Sec.~\ref{supp:failure_breakdown}).

Building upon these observations, we propose DA-CoT (Difficulty-Aware Chain-of-Thought), a training-free reasoning segmentation framework that adaptively triggers CoT reasoning based on the intrinsic difficulty of each sample, as estimated from the model’s attention maps.
As illustrated in Figure~\ref{fig:attention_map}, CoT-beneficial and CoT-detrimental cases exhibit markedly different attention patterns.
Interestingly, when CoT is unnecessary, the attention map from the $<$SOS$>$ token to the visual tokens (hereafter referred to as the \textit{attention map}) shows low entropy, indicating that the model’s focus is already well localized to the relevant region.
In contrast, CoT-beneficial cases exhibit high-entropy attention, reflecting dispersed focus and uncertainty in spatial grounding.
These findings suggest that the model’s internal attention entropy serves as a reliable signal for determining whether CoT reasoning should be applied.
Leveraging this signal, DA-CoT effectively distinguishes cases that do not require CoT, thus avoiding overthinking. Consequently, by adaptively bypassing unnecessary reasoning, it significantly improves inference efficiency.

Furthermore, to enhance the robustness of the proposed adaptive reasoning path, we introduce a \textit{Bidirectional Rethinking Mechanism}, which provides a fallback strategy when the initially selected reasoning path fails to produce a valid output.
In such cases, the model automatically reverts to the alternative reasoning mode (with or without CoT), mimicking the human tendency to “rethink” a problem from the opposite perspective.
This dual-path reasoning grants the model a second opportunity to recover from potential failures caused by either overthinking or underthinking.

Extensive experiments on Reasoning Segmentation and Referring Segmentation benchmarks demonstrate that our approach consistently improves performance, yielding strong results overall and demonstrating superior performance compared to prior works on Reasoning Segmentation.
Interestingly, DA-CoT outperforms existing methods that enhance their CoT capabilities through Reinforcement Learning~\cite{segzero, samr1, visionreasoner}, highlighting the strength of our training-free adaptive design.
Moreover, when integrated into existing learning-based reasoning frameworks, our method consistently enhances their performance, confirming its compatibility and general applicability.
Comprehensive ablation studies further validate the effectiveness of each proposed component.

\noindent We make the following contributions:
\begin{itemize}

\item We challenge the prevailing view that Chain-of-Thought (CoT) prompting should be applied by default, revealing that indiscriminate CoT usage can cause \textit{overthinking} and degrade grounding accuracy in reasoning segmentation.

\item We discover that the entropy of \textit{attention maps} reflects task difficulty, providing a simple yet effective intrinsic signal for determining when CoT reasoning is beneficial or detrimental.

\item We propose a \textbf{training-free} adaptive reasoning framework that integrates \textbf{DA-CoT} (Difficulty-Aware Chain-of-Thought) and a \textbf{Bidirectional Rethinking Mechanism}, where DA-CoT selectively applies CoT reasoning based on attention entropy to avoid overthinking, and the rethinking mechanism enhances robustness by enabling automatic fallback between reasoning modes for recovering from both overthinking- and underthinking-induced failures.

\item Our approach outperforms prior training-based methods on ReasonSeg, while achieving competitive performance on Referring Segmentation. Furthermore, it demonstrates lower inference time and plug-and-play compatibility across diverse MLLM scales and architectures.

\end{itemize}

\section{Related Work}
\label{sec:related_work}

\subsection{Multimodal Large Language Models (MLLMs)}
Building on the success of LLMs, early MLLMs such as LLaVA~\cite{llava,
llava15} and BLIP-2~\cite{blip2} aligned pre-trained vision encoders with
LLMs through visual instruction tuning. Recent models, including the
Qwen-VL series~\cite{qwen25vl} and GLM-4.1V-Thinking~\cite{GLM41},
further improve multimodal reasoning through larger scales and
reinforcement learning. These advances enable complex tasks beyond VQA,
including the reasoning segmentation studied in this work.

\subsection{Reasoning Segmentation}
\vspace{-1mm}
The task of Reasoning Segmentation was first proposed by LISA~\cite{lisa}, requiring models to generate segmentation masks from complex and implicit text queries. LISA pioneered the ``embedding-as-mask'' paradigm, adding a $<$SEG$>$ token to the MLLM's vocabulary to grant it pixel-level generation capabilities.

Subsequent works refine this paradigm for efficiency, replacing the
external segmentation model with a lightweight decoder~\cite{pixellm} or
prompting SAM~\cite{sam} with MLLM-generated points~\cite{sam4mllm}.
In a different direction, RSVP~\cite{rsvp} presented a training-free two-stage framework separating reasoning from segmentation, leveraging visual prompting to utilize the MLLM's inherent localization abilities. Also training-free, ThinkFirst~\cite{thinkfirst} prompts GPT-4o to produce a chain-of-thought description of the image that conditions a downstream segmentation assistant, applying CoT unconditionally to every query.

Advanced training strategies have also been a key research focus. Seg-Zero~\cite{segzero} demonstrated that reasoning abilities can be elicited via cognitive reinforcement learning (RL) without explicit reasoning data, achieving strong zero-shot performance on the ReasonSeg benchmark. SAM-R1~\cite{samr1} also utilized RL, uniquely integrating SAM 2~\cite{sam2} as a reward provider to enhance alignment between reasoning and segmentation. Recently, SAM-Veteran~\cite{sam-veteran} extended this paradigm by training an MLLM agent to mimic interactive segmentation through iterative mask refinement.

While these training-based methods have successfully advanced the field, they all apply a fixed reasoning strategy. Our work departs from this training-centric approach to propose a training-free framework that dynamically modulates the reasoning strategy.

\subsection{Chain-of-Thought in Multimodal Contexts}
Chain-of-Thought was a prompting technique proposed for LLMs to solve
complex reasoning problems by generating intermediate steps~\cite{cot}.
This method has been adapted for MLLMs. Models like
VisionReasoner~\cite{visionreasoner} have sought to improve performance
by generating structured reasoning processes, and other studies have
explored integrating visual information into the
CoT~\cite{multimodalcot, visualcot}.

While such efforts to enhance CoT capabilities have been active, the question of when to activate CoT has remained unaddressed. CoT is a powerful tool, but its benefits are not universal.
Recent studies in NLP have pointed out that applying CoT to simple problems that do not require multi-step reasoning can induce ``overthinking,'' leading to performance degradation~\cite{adaptive_thought, cotornot}.
This risk is acute in reasoning segmentation, where unnecessary CoT can
disrupt the precise visual grounding the task depends on.
Tian et al.~\cite{vapo} recently showed that in reasoning-trained MLLMs, prolonged reasoning progressively weakens perceptual grounding, and mitigated this effect by retraining the model with vision-anchored policy optimization.

In contrast to existing research focused on enhancing CoT capabilities through training~\cite{vapo, samr1, segzero, visionreasoner}, our work explores the problem of when to activate CoT. We analyze the model's internal attention mechanisms to propose an adaptive, training-free framework that selectively leverages the strengths of CoT while avoiding its potential pitfalls.

\vspace{-2mm}
\section{Preliminary Analysis}
\label{sec:preliminary_analysis}
\vspace{-1mm}
We observe, through empirical findings such as Figure~\ref{fig:attention_map}, that it may be possible to determine whether CoT prompting should be applied to a given sample based on \textit{intrinsic signals}—without requiring any additional training.
This section presents preliminary analyses aimed at establishing such signals as quantifiable indicators for CoT triggering.

\begin{figure*}[t]
\centering
\includegraphics[width=1.0\linewidth]{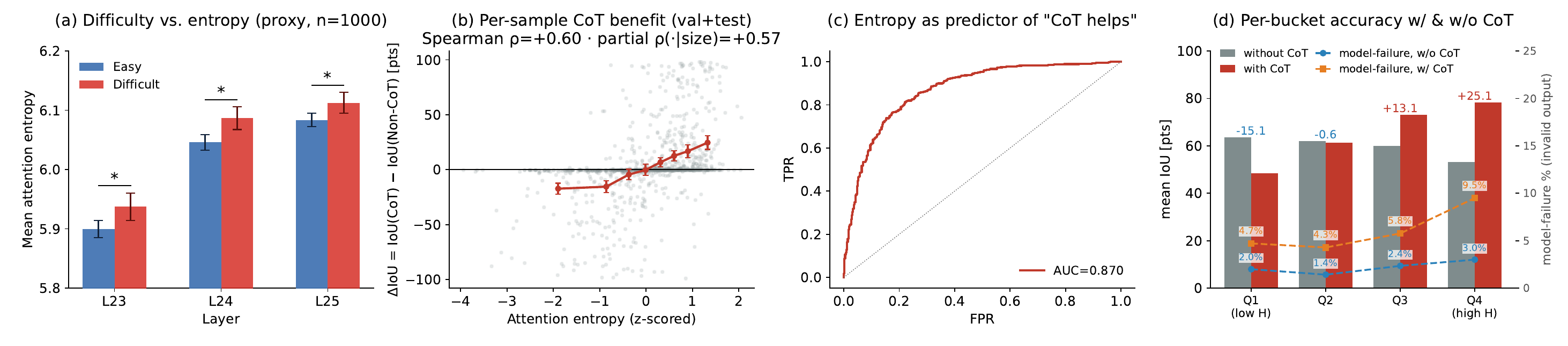}
\vspace{-8mm}
  \caption{
  Attention entropy predicts sample difficulty and per-sample CoT benefit using Qwen2.5VL-7B.
  (a)~Mean attention entropy of easy vs.\ difficult samples on the proxy set ($n=1{,}000$) at the three layers selected in Sec.~\ref{sec:method}; error bars: $\pm 1$ SEM; *: Mann--Whitney $U$ $p<0.05$.
  (b)~Per-sample $\Delta$IoU vs.\ attention entropy (z-scored within each split) on held-out ReasonSeg val+test (invalid outputs score IoU $= 0$); gray: individual samples, red: octile means with 95\% CI.
  (c)~ROC for predicting $\Delta\text{IoU} > 0$ from entropy alone (ties excluded).
  (d)~Mean IoU per entropy quartile with and without CoT, computed over each mode's valid outputs (bars, left axis), and per-mode model-failure rates (dashed, right axis; Sec.~\ref{supp:failure_detection}).}
\label{fig:mean_entropy}
\vspace{-6mm}
\end{figure*}

\vspace{-1mm}
\subsection{Entropy of Attention Map}
\label{sec:entropy}
\vspace{-1mm}
We begin by formalizing the entropy of the attention map as an intrinsic signal for reasoning behavior in the reasoning segmentation task.
In this setting, the input to the multi-modal model consists of a query sentence, an image, and a special \texttt{<SOS>} token.
The full input sequence includes both language tokens and visual tokens, which are jointly encoded by the transformer backbone.

To isolate the model's grounding behavior, we examine the attention map from the \texttt{<SOS>} token to only the visual tokens.
Formally, let $Q_{sos} \in \mathbb{R}^{1 \times d}$ denote the query vector corresponding to the \texttt{<SOS>} token, and let $K_{vis} \in \mathbb{R}^{N_\text{vis} \times d}$ represent the key matrix of $N$ visual tokens extracted from the image, where $d$ is the hidden dimension.

For each attention head $h \in \{1, ..., H\}$, the attention weights are computed as $\mathbf{a}_h = \text{softmax} \left( \frac{Q_{sos}^h (K_{vis}^h)^\top}{\sqrt{d/H}} \right)$, where $\mathbf{a}_h \in \mathbb{R}^{N_\text{vis}}$ is the attention distribution over the visual tokens for head $h$. To obtain the final attention map, we average the attention weights across all heads: $\mathbf{a} = \frac{1}{H} \sum_{h=1}^{H} \mathbf{a}_h$. Using Shannon entropy, we define the \textit{attention entropy} as $\mathcal{H}_{\text{attn}} = - \sum_{i=1}^{N_\text{vis}} \mathbf{a}[i] \log \mathbf{a}[i]$, which quantifies the dispersion of attention over the visual tokens as a single scalar value.

A low entropy value indicates that the model's attention is sharply focused on specific visual regions, suggesting confident grounding, whereas high entropy reflects a more diffuse focus, typically associated with uncertainty in resolving the query.
Intuitively, the \texttt{<SOS>} token reflects the model's grounding state before reasoning begins: when the answer is directly resolvable, attention concentrates on the target, yielding low entropy; when reasoning is required, the answer is not yet resolved, leaving attention dispersed.
This metric offers a lightweight and training-free estimate of the model’s internal uncertainty regarding visual grounding. Further analysis of this entropy formulation is provided in the supplementary material (Sec.~\ref{supp:entropy_validity}).

\vspace{-1mm}
\subsection{Difficulty Awareness via Entropy}
\label{sec:difficulty}
Attention entropy, as defined in Section~\ref{sec:entropy}, quantifies how concentrated the model's visual focus is.
Following Kang et al.~\cite{kang}, we sample 1,000 instances from the training splits of segmentation datasets~\cite{refcoco, res}, run inference with Qwen2.5VL-7B without CoT prompting, and label samples with IoU $\leq 0.5$ as \textit{difficult} and the rest as \textit{easy}.
Across the three layers later selected by our variance criterion (layers 23--25; Sec.~\ref{sec:method}), difficult samples exhibit significantly higher entropy (Fig.~\ref{fig:mean_entropy}(a); Mann--Whitney $U$, $p<0.05$).
Since our framework triggers CoT per sample (Sec.~\ref{sec:method}), we further examine the effect of CoT at the level of individual samples.
We run inference twice, with and without CoT prompting, on all held-out ReasonSeg val/test samples and measure $\Delta\text{IoU} = \text{IoU}_{\text{CoT}} - \text{IoU}_{\text{Non-CoT}}$.
Entropy, aggregated over the same three layers, shows a monotone relation with this benefit (Fig.~\ref{fig:mean_entropy}(b); Spearman $\rho = +0.60$, and $+0.57$ after partialling out log target size, Sec.~\ref{supp:per_sample}): CoT degrades low-entropy samples and improves high-entropy ones.
Used as a per-sample predictor of whether CoT helps ($\Delta\text{IoU} > 0$), entropy attains AUC $= 0.870$ (Fig.~\ref{fig:mean_entropy}(c)).
Fig.~\ref{fig:mean_entropy}(d) further shows an entropy-dependent trend that holds despite higher failure rates: CoT degrades low-entropy samples, while becoming increasingly beneficial as entropy rises. Model failures are more frequent under CoT, motivating the Bidirectional Rethinking mechanism to recover from such cases (Secs.~\ref{sec:rethinking}, \ref{supp:sec:biret}).

\section{Method}
\label{sec:method}

Building upon Section~\ref{sec:preliminary_analysis}, we propose DA-CoT, a training-free framework designed to mitigate \emph{overthinking} in reasoning segmentation. DA-CoT adaptively determines whether to apply CoT prompting by leveraging the intrinsic attention-entropy signal, and further enhances robustness through a human-inspired rethinking mechanism.

\subsection{Overview of DA-CoT framework}
\label{sec:overview}

\begin{figure}[t]
\centering
\includegraphics[width=1.0\linewidth]{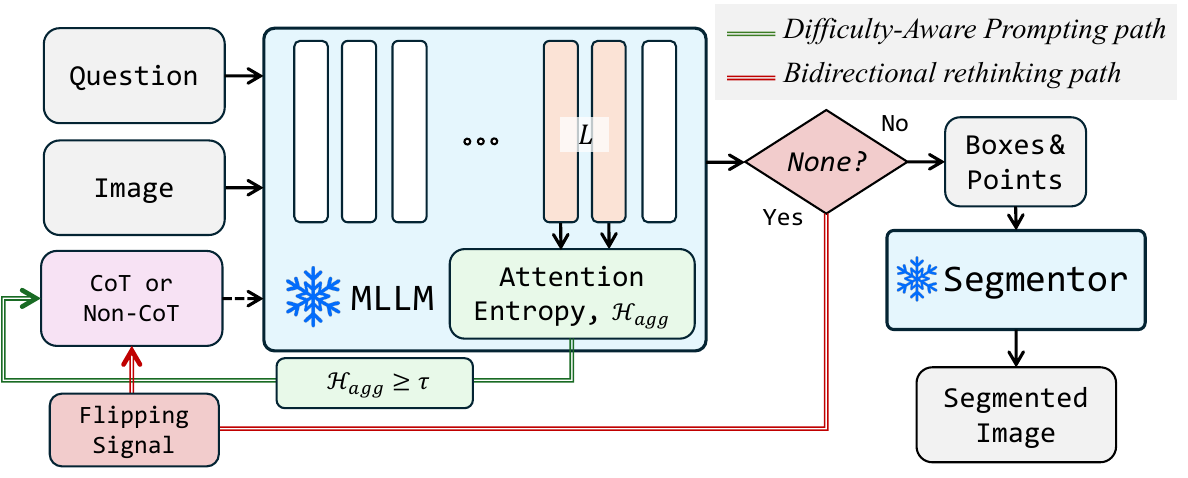}
\vspace{-6mm}
\caption{
Overview of the proposed \textbf{DA-CoT} framework during inference.
The aggregated attention entropy $\mathcal{H}_{\text{agg}}$, computed over the pre-identified layers $\mathcal{L}$, is used to adaptively decide whether to trigger CoT reasoning.
If $\mathcal{H}_{\text{agg}} \ge \tau$, the model follows the \textbf{Difficulty-Aware Prompting Path} with CoT for structured reasoning; otherwise, it takes the direct path without CoT for efficient inference.
When the initial reasoning path fails to produce a valid output, a discrete \textit{flipping signal} activates the \textbf{Bidirectional Rethinking Path}, enabling re-evaluation and recovery from overthinking or underthinking.
The final bounding boxes or point prompts are then passed to the segmentation head to produce the reasoning-aware segmentation result.
}

\label{fig:main_fig}
\vspace{-5mm}
\end{figure}

Figure~\ref{fig:main_fig} presents the overall structure of the proposed DA-CoT framework, which operates in three stages:
(1) layer and threshold determination,
(2) difficulty-aware CoT, and
(3) bidirectional rethinking.

\noindent\textbf{Layer and Threshold Determination Phase.}
Using a fixed proxy set $\mathcal{D}_{\text{proxy}}$, we identify the top-$k$ layers
$\mathcal{L}$ exhibiting the largest attention-entropy variance across samples.
From the aggregated entropy distribution of these layers, the second elbow point is selected as the attention threshold $\tau$ (Sec.~\ref{sec:calibration}). These parameters are fixed for the backbone model and directly applied across downstream datasets.

\noindent\textbf{Difficulty-Aware CoT Phase.}
Given an image--query pair, the model computes the aggregated attention entropy $\mathcal{H}_{\text{agg}}$ using the pre-selected layers $\mathcal{L}$.
If $\mathcal{H}_{\text{agg}}$ exceeds the threshold $\tau$, the model activates CoT prompting to engage in structured reasoning;
otherwise, it proceeds with direct prediction without additional reasoning steps.
This selective mechanism prevents unnecessary reasoning on easy samples while maintaining high reasoning capability for ambiguous ones.

\noindent\textbf{Bidirectional Rethinking Phase.}
When the initial reasoning path fails to yield a valid output, a discrete flipping signal reverses the reasoning decision (activating CoT if it was disabled, or vice versa) and triggers
a single second inference pass (Sec.~\ref{sec:rethinking}). Finally, the MLLM outputs bounding boxes or point prompts, which are passed to a segmentation head to generate the final reasoning-aware segmentation result.

\subsection{Determining Attention Layers \& Threshold}
\label{sec:calibration}

To enable adaptive reasoning, DA-CoT relies solely on the internal attention entropy extracted from pretrained MLLMs.
We design a procedure to determine two key components:
(i) the set of layers $\mathcal{L}$ most responsive to reasoning difficulty, and
(ii) the attention threshold $\tau$ for CoT triggering.
These parameters are computed once for a given backbone and then reused across downstream benchmarks.

\noindent\textbf{Proxy Set.}
We construct a proxy dataset $\mathcal{D}_{\text{proxy}}$ consisting of 1{,}000 image--query pairs randomly sampled from the \textbf{training splits} of segmentation datasets~\cite{refcoco, res}.
This procedure requires \emph{no segmentation labels}; we only run a standard forward pass (without CoT) to extract attention maps and compute entropies. Further analysis of the proxy set is provided in the supplementary material (Sec.~\ref{supp:proxy_robustness}, Sec.~\ref{supp:proxy_free}).

\noindent\textbf{Layer Selection.}
The layers for entropy measurement should satisfy two implicit conditions: (1) being highly correlated with the model’s grounding capability, and (2) being sensitive to differences in reasoning difficulty.
Although these properties cannot be directly observed, they can be approximated by identifying layers with high entropy variance across the proxy set as $\mathcal{L} = \operatorname{Topk}_{\ell}\big(\operatorname{Var}_{x \in \mathcal{D}_{\text{proxy}}} [\mathcal{H}_{\text{attn}}^{(\ell)}(x)] \big)$.
Layers exhibiting large variance effectively capture the transition between confident (low-entropy) and uncertain (high-entropy) grounding behaviors, thus serving as discriminative indicators of difficulty.
In our experiments, $k=3$ layers provide stable and generalizable performance. Notably, this simple variance-based approach consistently identifies mid-to-late layers, aligning with recent findings~\cite{kang, mllm_grounding_layer2, how} that highlight these layers as significant for multi-modal reasoning and grounding.

\begin{figure}[t]
\centering
\includegraphics[width=1.0\linewidth]{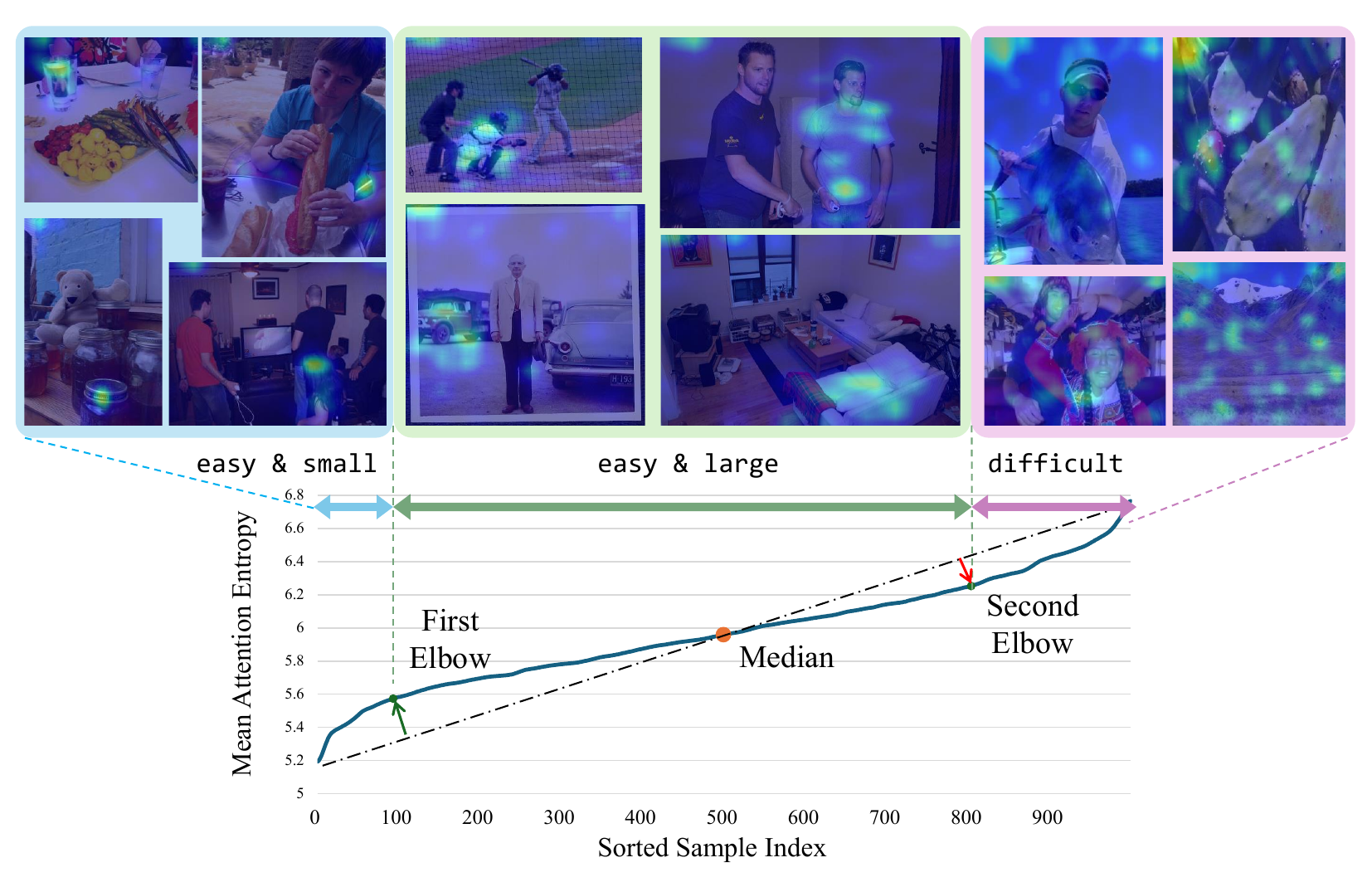}
\vspace{-8mm}
\caption{
Distribution of aggregated attention entropy $\mathcal{H}_{\text{agg}}$ over the proxy set.
The curve exhibits an S-shaped trend with two elbow points that divide samples by reasoning difficulty.
Samples below the first elbow correspond to easy and small-object cases with confident grounding,
those between elbows represent medium-complexity examples with larger objects,
and those beyond the second elbow correspond to difficult reasoning cases that require CoT prompting.
The second elbow is selected as the decision threshold $\tau$ for adaptive CoT activation.
}
\label{fig:secondary_elbow_vis}
\vspace{-2mm}
\end{figure}

\noindent\textbf{Attention Threshold.}
For each sample $x_i \in \mathcal{D}_{\text{proxy}}$, the aggregated attention entropy is defined as $\mathcal{H}_{\text{agg}}(x_i) = \frac{1}{k} \sum_{\ell \in \mathcal{L}} \mathcal{H}_{\text{attn}}^{(\ell)}(x_i)$.
Sorting $\mathcal{H}_{\text{agg}}$ values in ascending order yields an S-shaped distribution (Figure~\ref{fig:secondary_elbow_vis}).
To identify elbow points, we employ the perpendicular-distance--based elbow detection method~\cite{kneedle_algorithm}: we first find the point with the maximum perpendicular distance to the line connecting the first and last points (first elbow), and then apply the same procedure to the tail subsequence to obtain the second elbow. Detailed mathematical formulations are provided in the supplementary material (Sec.~\ref{supp:sec:elbow}).
We empirically determine the second elbow point as the decision threshold $\tau$ given by $\tau = \text{SecondElbow}\big(\{\mathcal{H}_{\text{agg}}(x_i)\}_{i=1}^{|\mathcal{D}_{\text{proxy}}|}\big)$.
As described in Figure~\ref{fig:secondary_elbow_vis}, samples below the first elbow correspond to easy and small-object cases, those between elbows to medium-complexity samples with relatively larger objects, and those beyond the second elbow to difficult reasoning cases requiring CoT prompting. While this stratification partly tracks target size, a size-controlled analysis shows that entropy predicts difficulty and CoT benefit beyond size alone (Sec.~\ref{supp:per_sample}).

Using $\mathcal{D}_{\text{proxy}}$, $\tau$ activates CoT in only about 21.2\% of cases---achieving superior efficiency compared to the first elbow point that would invoke CoT in 90.8\% of cases, as shown in Table~\ref{tab:entropy_threshold}.

\begin{table}[t]
\centering
\caption{Comparative analysis of entropy threshold selection strategies. Performance (gIoU in \%) and CoT usage are measured without Bidirectional Rethinking.  Evaluation is performed on the proxy set described in Section~\ref{sec:difficulty}.}
\label{tab:entropy_threshold}
\vspace{-2mm}
\resizebox{\columnwidth}{!}{
\begin{tabular}{l|ccccc}
\toprule
Threshold Strategy & All-CoT & First Elbow & Median & Second Elbow & All-Non-CoT \\
\midrule
gIoU & 66.7 & 67.3 & 67.8 \scriptsize & 68.2 \scriptsize & 62.1 \\
CoT Usage (\%) & 100.0 & 90.8 & 49.2 \scriptsize & 21.2 \scriptsize & 0.0 \\
\bottomrule
\end{tabular}
}
\vspace{-0.4cm}
\end{table}

\subsection{Bidirectional Rethinking}
\label{sec:rethinking}

While DA-CoT effectively adapts CoT prompting using intrinsic signals, reasoning segmentation may still involve ambiguous visual-linguistic grounding where deterministic triggering is insufficient. To strengthen robustness, we introduce a Bidirectional Rethinking mechanism, inspired by how humans reassess an initial judgment upon sensing that a problem was oversimplified or overcomplicated.

When the initial reasoning attempt produces incomplete or invalid outputs (\eg missing coordinates or coordinate fields), DA-CoT automatically performs a rethinking step.
In this step, the model reprocesses the same input through the complementary reasoning path, providing a rule-based fallback that enhances output reliability.
This mechanism grants the model a second opportunity to recover from failures caused by excessive or insufficient reasoning, thereby reinforcing the robustness of the framework.

\section{Experiments}
We use generalized Intersection-over-Union~(gIoU) for reasoning segmentation and cumulative Intersection-over-Union~(cIoU) for referring segmentation as evaluation metrics, following prior work~\cite{lisa, visionreasoner, sam-veteran, samr1, segzero}.
Best and second-best scores are shown in \textbf{bold} and \underline{underlined}, respectively, and our method is highlighted in \colorbox[HTML]{e0e0e0}{gray}.

\subsection{Experimental Settings}
\noindent \textbf{Datasets.} We evaluate our method on a comprehensive set of standard benchmarks for reasoning and referring expression segmentation. For reasoning segmentation, we use the ReasonSeg dataset~\cite{lisa}, which is specifically designed to test complex visual-linguistic reasoning abilities. For referring expression segmentation, we use the widely-adopted RefCOCO, RefCOCO+, and RefCOCOg datasets~\cite{refcoco, res}. Following established protocols, we use the official test splits for evaluation.

\noindent \textbf{Implementation Details.}
Our experiments are primarily conducted on the Qwen2.5VL-7B model. Accordingly, unless stated otherwise, the results for our proposed DA-CoT-7B method in the tables refer to this Qwen2.5VL-7B-based implementation. For mask generation, we employ SAM2-Large~\cite{sam2}, which takes positional prompts from the MLLM. To demonstrate the plug-and-play generalizability of our method, we also test it on a diverse set of models, including Seg-Zero, VisionReasoner, Qwen2.5VL-32B, Qwen2.5VL-72B, Qwen3VL-8B, and GLM4.1V-9B. For image preprocessing, inputs for the Qwen models are resized to 840 $\times$ 840 pixels. Following prior work~\cite{lisa, segzero, visionreasoner, samr1, sam-veteran}, all methods use deterministic decoding, and we report single-run results. The complete procedure is detailed in the supplementary material (Sec.~\ref{supp:dacot_algorithm}).

\subsection{Main Results}
We evaluate our proposed method on two challenging tasks: Reasoning Segmentation and Referring Expression Segmentation. The result for Seg-Zero-7B$^{\dagger}$ was re-evaluated in SAM-R1~\cite{samr1} using the officially released model weights.
\begin{table}[t]
\centering
\caption{Performance comparison with zero-shot models on ReasonSeg benchmark. Performance is reported as gIoU.
{$^{\dagger}$ denotes performance re-evaluated by~\cite{samr1}}.}
\label{tab:reasonseg}
\vspace{-0.7mm}
\renewcommand{\arraystretch}{0.85}
\resizebox{0.8\linewidth}{!}{
\begin{tabular}{l|ccc}
\toprule
\multicolumn{1}{c|}{\multirow{2}{*}{Method}} & \multicolumn{3}{c}{ReasonSeg} \\
\cmidrule(lr){2-4}
 & val & test & Avg. \\
\midrule
\multicolumn{4}{l}{\textit{Training-based methods}} \\
\midrule
OVSeg~\cite{ovseg} & 28.5 & 26.1 & 27.3 \\
ReLA~\cite{rela} & 22.4 & 21.3 & 21.9 \\
SAM4MLLM~\cite{sam4mllm} & 46.7 & - & - \\
LISA-7B-LLaVA1.5~\cite{lisa} & 53.6 & 48.7 & 51.2 \\
LISA-13B-LLaVA1.5~\cite{lisa} & 57.7 & 53.8 & 55.8 \\
Seg-Zero-7B$^{\dagger}$~\cite{segzero} & 62.0 & 58.3 & 60.2 \\
SAM-R1-7B~\cite{samr1} & 64.0 & 60.2 & 62.1 \\
VisionReasoner-7B~\cite{visionreasoner} & \underline{66.3} & \textbf{63.6} & \underline{65.0} \\
SAM-Veteran-7B~\cite{sam-veteran} & \textbf{68.2} & \underline{62.6} & \textbf{65.4} \\
\midrule
\multicolumn{4}{l}{\textit{Training-free methods}} \\
\midrule
Grounded-SAM~\cite{groundedsam} & 26.0 & 21.3 & 23.7 \\
ThinkFirst~\cite{thinkfirst} & - & 48.6 & - \\
RSVP-Qwen~\cite{rsvp} & 58.6 & 56.6 & 57.6 \\
SAM3 Agent $+$ Qwen-7B~\cite{sam3} & \underline{62.2} & \underline{63.0} & \underline{62.6} \\
\rowcolor[HTML]{e0e0e0} DA-CoT-7B & \textbf{68.0} & \textbf{64.6} & \textbf{66.3} \\
\bottomrule
\end{tabular}
}
\vspace{-4mm}
\end{table}

\noindent \textbf{Reasoning Segmentation.}
We first compare our method against existing models on the ReasonSeg benchmark.
As presented in Table~\ref{tab:reasonseg}, our method achieves state-of-the-art performance on the ReasonSeg Benchmark. Remarkably, our training-free approach outperforms even task-specifically trained systems, including the strongest RL-trained methods~\cite{segzero, samr1, visionreasoner, sam-veteran} that share our Qwen2.5-VL-7B backbone and SAM2-based mask generation.

\noindent \textbf{Referring Expression Segmentation.}
To assess the generality of our method beyond complex reasoning tasks, we evaluate it on standard referring benchmarks. We use the testA set for RefCOCO and RefCOCO+, and the test set for RefCOCOg.

As shown in Table~\ref{tab:referring_seg}, DA-CoT also achieves competitive performance on standard referring expression segmentation, indicating that its benefits are not confined to reasoning-intensive queries.

\noindent \textbf{Broader Vision-Language Tasks.}
To further assess the applicability of DA-CoT beyond segmentation, we evaluate our method on two broader vision-language benchmarks, LogicVista~\cite{logicvista} and MMVP~\cite{mmvp}. LogicVista evaluates multimodal logical reasoning in diverse visual contexts, while MMVP assesses fine-grained visual understanding through challenging image pairs with subtle visual differences. As shown in Table~\ref{tab:dacot_reasoning_comparison}, DA-CoT consistently improves over All-CoT and All-Non-CoT on both benchmarks, demonstrating the broader applicability of our adaptive reasoning strategy.

\subsection{Plug-and-Play Generalization}
A key advantage of our method is its potential for plug-and-play generalization. We investigate this capability across both different reasoning methods and different baseline MLLMs.

\begin{table}[t]
\centering
\caption{Performance comparison on Referring Expression Segmentation benchmark. Performance is reported as cIoU. {$^{\dagger}$ denotes performance re-evaluated by~\cite{samr1}}.}
\label{tab:referring_seg}

\renewcommand{\arraystretch}{0.85}
\resizebox{0.8\linewidth}{!}{
\begin{tabular}{l|cccc}
\toprule
\multicolumn{1}{c|}{\multirow{2}{*}{Method}} & \multicolumn{4}{c}{ReferringSeg} \\
\cmidrule(lr){2-5}
 & COCO & COCO+ & COCOg & Avg. \\
\midrule
\multicolumn{5}{l}{\textit{Training-based methods}} \\
\midrule
LAVT~\cite{lavt} & 75.8 & 68.4 & 62.1 & 68.8 \\
ReLA~\cite{rela} & 76.5 & 71.0 & 66.0 & 71.2 \\
LISA-7B-LLaVA1.5~\cite{lisa} & 76.5 & 67.4 & 68.5 & 70.8 \\
PixelLM-7B~\cite{pixellm} & 76.5 & 71.7 & 70.5 & 72.9 \\
PerceptionGPT-7B~\cite{perceptiongpt} & 78.6 & 73.9 & 71.7 & 74.7 \\
VisionReasoner-7B~\cite{visionreasoner} & 78.9 & \underline{74.9} & 71.3 & 75.0 \\
Seg-Zero-7B$^{\dagger}$~\cite{segzero} & \underline{79.2} & 73.9 & \underline{73.3} & 75.5 \\
SAM-R1-7B~\cite{samr1} & \underline{79.2} & 74.7 & 73.1 & \underline{75.7} \\
SAM-Veteran-7B~\cite{sam-veteran} & \textbf{80.8} & \textbf{76.6} & \textbf{73.4} & \textbf{76.9} \\
\midrule
\multicolumn{5}{l}{\textit{Training-free methods}} \\
\midrule
Yu et al.~\cite{Yu_et_al} & 23.6 & 24.9 & 31.1 & 26.5 \\
TAS~\cite{TAS} & 30.3 & 38.8 & 36.2 & 35.1 \\
Ref-Diff~\cite{Ref-Diff} & 37.4 & 38.7 & 38.0 & 38.0 \\
IteRPrimE~\cite{IteRPrimE} & 46.5 & 51.6 & 45.8 & 48.0 \\
Kang et al.~\cite{kang} & \textbf{78.9} & \underline{67.1} & \underline{69.0} & \underline{71.7} \\
\rowcolor[HTML]{e0e0e0} DA-CoT-7B & \underline{77.7} & \textbf{72.4} & \textbf{70.7} & \textbf{73.6} \\
\bottomrule
\end{tabular}
}
\vspace{-4mm}
\end{table}

\noindent \textbf{Generalization across MLLMs.}
To demonstrate that our framework's benefits are not confined to this specific baseline, we assessed its performance across a spectrum of more powerful models. This includes Seg-Zero and VisionReasoner which enhance CoT reasoning capabilities based on Qwen2.5VL-7B~\cite{qwen25vl}, larger-scale models like Qwen2.5VL-32B~\cite{qwen25vl} and Qwen2.5VL-72B~\cite{qwen25vl}, as well as the architecturally advanced Qwen3VL-8B~\cite{qwen3}, and GLM4.1V-9B~\cite{GLM41}. As shown in Table~\ref{tab:plugandplay_scalability}, our method consistently delivers performance gains across these models. This finding suggests that our approach is not narrowly tuned to a specific baseline and can effectively scale to enhance the performance of larger and more recent MLLMs.

\subsection{Ablation Studies}
\label{sec:ablation}
\textbf{Impact of Proposed Components.}
We conduct an ablation study to analyze the contribution of each component. First, we design an improved CoT prompt inspired by existing research~\cite{cot_motivation}, the details of which are provided in the supplementary material (Sec.~\ref{supp:sec:prompt}). Our baseline is established as All-CoT, where this improved CoT prompt is applied to all samples. This baseline already achieves competitive performance, providing a strong starting point for our experiments.

As shown in Table~\ref{tab:ablation}, applying our DA-CoT, which dynamically selects reasoning paths, provides a consistent performance improvement across both ReasonSeg and ReferringSeg benchmarks. Furthermore, the addition of Bidirectional Rethinking yields additional gains by enhancing the framework's robustness. This highlights its effectiveness in resolving initial failures regardless of the specific task.

\noindent \textbf{Inference Latency Comparison.} The reported inference time is measured end-to-end, encompassing all processes required for DA-CoT (e.g., attention map extraction and BiReT). All measurements were conducted on a single NVIDIA RTX 4090 GPU. As shown in Table~\ref{tab:efficiency}, DA-CoT significantly reduces latency compared to the All-CoT baseline while achieving higher performance. Furthermore, it demonstrates superior inference efficiency and overall performance compared to existing methods~\cite{lisa, segzero, visionreasoner}.
\begin{table}[t]
\centering
\caption{Performance comparison on LogicVista and MMVP. We report accuracy for LogicVista and pair accuracy for MMVP.}
\label{tab:dacot_reasoning_comparison}
\renewcommand{\arraystretch}{0.85}
\resizebox{0.8\linewidth}{!}{
\begin{tabular}{l|cc}
\toprule
\multicolumn{1}{c|}{Method}
& LogicVista
& MMVP \\
\midrule
Qwen2.5VL-7B~\cite{qwen25vl} (All-Non-CoT)
& 37.81
& 58.00 \\
Qwen2.5VL-7B~\cite{qwen25vl} (All-CoT)
& \underline{44.30}
& 55.33 \\
VisionReasoner-7B~\cite{visionreasoner}
& 38.93
& 56.00 \\
Seg-Zero-7B~\cite{segzero}
& 38.93
& \underline{58.67} \\
\rowcolor[HTML]{e0e0e0} DA-CoT-7B
& \textbf{45.64}
& \textbf{59.33} \\
\bottomrule
\end{tabular}
}
\end{table}

\begin{table}[t]
\centering
\caption{Scalability and robustness of the proposed method on larger and more recent MLLMs. Performance is reported as gIoU. $^{\dagger}$ denotes performance re-evaluated by~\cite{samr1}.}
\label{tab:plugandplay_scalability}
\vspace{-2.4mm}
\renewcommand{\arraystretch}{0.85}
\resizebox{0.8\linewidth}{!}{
\begin{tabular}{l|ccc}
\toprule
\multicolumn{1}{c|}{\multirow{2}{*}{Method}} & \multicolumn{3}{c}{ReasonSeg} \\
\cmidrule(lr){2-4}
 & val & test & Avg. \\
\midrule
Seg-Zero-7B$^{\dagger}$~\cite{segzero} & 62.0 & 58.3 & 60.2 \\
\rowcolor[HTML]{e0e0e0} Seg-Zero-7B + Ours & \textbf{66.2} \scriptsize \textcolor{red}{($\uparrow$ 4.2)} & \textbf{64.6} \scriptsize \textcolor{red}{($\uparrow$ 6.3)} & \textbf{65.4} \scriptsize \textcolor{red}{($\uparrow$ 5.2)} \\
\midrule
VisionReasoner-7B~\cite{visionreasoner} & 66.3 & 63.6 & 65.0 \\
\rowcolor[HTML]{e0e0e0} VisionReasoner-7B + Ours & \textbf{67.4} \scriptsize \textcolor{red}{($\uparrow$ 1.1)} & \textbf{65.6} \scriptsize \textcolor{red}{($\uparrow$ 2.0)} & \textbf{66.5} \scriptsize \textcolor{red}{($\uparrow$ 1.5)} \\
\midrule
Qwen2.5VL-7B~\cite{qwen25vl} & 63.0 & 60.7 & 61.9 \\
\rowcolor[HTML]{e0e0e0} Qwen2.5VL-7B + Ours & \textbf{68.0} \scriptsize \textcolor{red}{($\uparrow$ 5.0)} & \textbf{64.6} \scriptsize \textcolor{red}{($\uparrow$ 3.9)} & \textbf{66.3} \scriptsize \textcolor{red}{($\uparrow$ 4.4)} \\
\midrule
Qwen2.5VL-32B~\cite{qwen25vl} & 63.1 & 62.5 & 62.8 \\
\rowcolor[HTML]{e0e0e0} Qwen2.5VL-32B + Ours & \textbf{68.1} \scriptsize \textcolor{red}{($\uparrow$ 5.0)} & \textbf{64.9} \scriptsize \textcolor{red}{($\uparrow$ 2.4)} & \textbf{66.5} \scriptsize \textcolor{red}{($\uparrow$ 3.7)} \\
\midrule
Qwen2.5VL-72B~\cite{qwen25vl} & 66.6 & 64.0 & 65.3 \\
\rowcolor[HTML]{e0e0e0} Qwen2.5VL-72B + Ours & \textbf{69.9} \scriptsize \textcolor{red}{($\uparrow$ 3.3)} & \textbf{68.0} \scriptsize \textcolor{red}{($\uparrow$ 4.0)} & \textbf{69.0} \scriptsize \textcolor{red}{($\uparrow$ 3.7)} \\
\midrule
Qwen3VL-8B~\cite{qwen3} & 67.9 & 67.4 & 67.7 \\
\rowcolor[HTML]{e0e0e0} Qwen3VL-8B + Ours & \textbf{69.2} \scriptsize \textcolor{red}{($\uparrow$ 1.3)} & \textbf{69.1} \scriptsize \textcolor{red}{($\uparrow$ 1.7)} & \textbf{69.2} \scriptsize \textcolor{red}{($\uparrow$ 1.5)} \\
\midrule
GLM4.1V-9B~\cite{GLM41} & 66.5 & 63.5 & 65.0 \\
\rowcolor[HTML]{e0e0e0} GLM4.1V-9B + Ours & \textbf{68.4} \scriptsize \textcolor{red}{($\uparrow$ 1.9)} & \textbf{66.7} \scriptsize \textcolor{red}{($\uparrow$ 3.2)} & \textbf{67.6} \scriptsize \textcolor{red}{($\uparrow$ 2.6)} \\
\bottomrule
\end{tabular}
}
\vspace{-4mm}
\end{table}

\noindent \textbf{Analysis of Thresholding Strategy.} We compared several thresholding strategies, as shown in Table~\ref{tab:entropy_threshold_reasonseg}. While adaptive CoT triggering proved more effective than both All-CoT and All-Non-CoT, our proposed second elbow strategy yielded the best performance. This result indicates that our approach finds an optimal balance between performance and efficiency, as it most effectively curtails unnecessary CoT usage while maximizing accuracy.

\noindent \textbf{Transfer Across Proxy Sources.}
The layers and threshold in all experiments are determined once on the default proxy and reused across every benchmark without re-determination.
To verify that this reuse does not depend on the proxy being drawn from segmentation data, we construct proxy sets of 1{,}000 image--query pairs from the validation splits of OK-VQA~\cite{okvqa} and GQA~\cite{gqa}, and the training split of A-OKVQA~\cite{aokvqa}, and apply the identical determination procedure of Sec.~\ref{sec:calibration}.
As shown in Table~\ref{tab:layer_threshold_transfer}, all sources select overlapping mid-to-late layers with thresholds $\tau \in [6.00, 6.21]$, and every configuration outperforms both All-CoT and All-Non-CoT on every benchmark.
The transfer also holds in the reverse direction: the Default row applies the segmentation-determined parameters unchanged to LogicVista and MMVP, matching Table~\ref{tab:dacot_reasoning_comparison}.
These results indicate that the parameters reflect an intrinsic property of the backbone rather than the source from which the proxy is drawn.

\noindent \textbf{Analysis of Aggregated Layers.} We also investigated the optimal number of layers ($k$) to aggregate for our difficulty signal. As shown in Table~\ref{tab:layer_ablation}, aggregating information from multiple layers led to more stable performance than relying on a single one, with both $k = 3$ and $k = 4$ showing competitive performance. However, $k=3$ was the optimal choice when considering the performance of the full model with Bidirectional Rethinking applied. We therefore adopted $k=3$ for all experiments.

\begin{table}[t]
\centering
\caption{
Ablation study on our proposed components:
DA-CoT and BiReT. The baseline applies the CoT prompt to all samples.
}
\label{tab:ablation}
\vspace{-3mm}
\renewcommand{\arraystretch}{0.85}
\resizebox{\linewidth}{!}{
\begin{tabular}{l|cc|ccc}
\toprule
\multicolumn{1}{c|}{\multirow{2}{*}{Method}} & \multicolumn{2}{c|}{ReasonSeg (gIoU)} & \multicolumn{3}{c}{ReferringSeg (cIoU)} \\
\cmidrule(lr){2-3} \cmidrule(lr){4-6}
 & val & test & RefCOCO & RefCOCO+ & RefCOCOg \\
\midrule
Qwen2.5VL-7B & 63.0 & 60.7 & 73.5 & 68.2 & 60.4 \\
+ DA-CoT
& 65.2 \scriptsize \textcolor{red}{($\uparrow$ 2.2)} & 64.0 \scriptsize \textcolor{red}{($\uparrow$ 3.3)} & 77.6 \scriptsize \textcolor{red}{($\uparrow$ 4.1)} & 72.0 \scriptsize \textcolor{red}{($\uparrow$ 3.8)} & 69.6 \scriptsize \textcolor{red}{($\uparrow$ 9.2)} \\
\rowcolor[HTML]{e0e0e0} +
BiReT
& \textbf{68.0} \scriptsize \textcolor{red}{($\uparrow$ 2.8)} & \textbf{64.6} \scriptsize \textcolor{red}{($\uparrow$ 0.6)} & \textbf{77.7} \scriptsize \textcolor{red}{($\uparrow$ 0.1)} & \textbf{72.4} \scriptsize \textcolor{red}{($\uparrow$ 0.4)} & \textbf{70.7} \scriptsize \textcolor{red}{($\uparrow$ 1.1)} \\
\bottomrule
\end{tabular}
}
\vspace{-4mm}
\end{table}

\begin{table}[t]
\centering
\caption{Inference latency and performance on the ReasonSeg test. All-CoT and All-Non-CoT denote applying CoT or direct prediction to all samples, respectively.}
\label{tab:efficiency}
\vspace{-2mm}
\renewcommand{\arraystretch}{0.85}
\resizebox{0.7\linewidth}{!}{
\begin{tabular}{l|cc}
    \toprule
    \multicolumn{1}{c|}{\textbf{Method}} & \textbf{time/img (sec)} & \textbf{gIoU (Test)} \\
    \midrule
    LISA-7B~\cite{lisa} & 7.0 & 48.7 \\
    Seg-Zero-7B~\cite{segzero} & 7.0 & 58.3 \\
    VisionReasoner-7B~\cite{visionreasoner} & 7.0 & 63.6 \\
    Qwen2.5VL-7B~\cite{qwen25vl} (All-Non-CoT) & \textbf{3.2} & 58.0 \\
    \rowcolor[HTML]{e0e0e0} Qwen2.5VL-7B~\cite{qwen25vl} (All-CoT) & 9.9 & 60.7 \\
    \rowcolor[HTML]{d0d0d0} DA-CoT-7B (w/o BiReT) & \underline{4.4} & \underline{64.0} \\
    \rowcolor[HTML]{c8c8c8} DA-CoT-7B (w/ BiReT) & {4.7} & \textbf{64.6} \\
    \bottomrule
\end{tabular}
}
\vspace{-3mm}
\end{table}

\begin{table}[t]
\centering
\caption{Ablation study on entropy threshold selection for ReasonSeg without Bidirectional Rethinking. Reported in gIoU.}
\label{tab:entropy_threshold_reasonseg}
\vspace{-2mm}
\renewcommand{\arraystretch}{0.85}
\resizebox{0.5\linewidth}{!}{
\begin{tabular}{l|cc|c}
\toprule
\multicolumn{1}{c|}{Threshold Strategy} & val & test & CoT Usage (\%) \\
\midrule
All-CoT & 63.0 & 60.7 & 100.0 \\
First Elbow & 63.9 & 61.0 & 76.1 \\
Median & 64.6 & 61.9 & 51.6 \\
\rowcolor[HTML]{e0e0e0} Second Elbow & \textbf{65.2} & \textbf{64.0} & \textbf{35.0} \\
All-Non-CoT & 59.9 & 58.0 & 0.0 \\
\bottomrule
\end{tabular}
}
\vspace{-4mm}
\end{table}

\noindent \textbf{Effectiveness of Strategy Flipping in BiReT.}
To further validate the design of BiReT, we analyze whether its performance gain stems merely from taking an additional inference attempt or from the explicit strategy flipping mechanism.
To this end, we compare BiReT against a fixed-strategy retry baseline, termed \textit{Unidirectional Rethinking} (\textit{UniReT}), which re-generates outputs without strategy switching (i.e., CoT $\rightarrow$ CoT or Non-CoT $\rightarrow$ Non-CoT). Following common evaluation practice, all methods in our experiments use \textit{deterministic decoding} unless otherwise noted, meaning that the same input under the same model and prompt yields a single fixed output. UniReT departs from this default only during the fallback phase by re-decoding the same reasoning mode under varying temperatures $t \in \{0.5, 0.7, 1.0\}$. Changing the temperature alters the token sampling distribution, so repeated runs on the same input can produce different reasoning traces and final predictions; we therefore report results averaged over 3 runs. We evaluate the effectiveness using the \textit{Valid Recovery Rate}, defined as the proportion of initially failed samples that successfully yield a valid response after the fallback phase, and the \textit{Correct Recovery Rate}, the proportion of initially failed samples whose recovered prediction achieves IoU $>$ 0.5.

As shown in Table~\ref{tab:biret_comparison}, UniReT peaks at 38.7\% Valid Recovery Rate and at most a 2.8\% Correct Recovery Rate, whereas BiReT achieves 97.2\% and 22.2\%, respectively, with the larger overall gain.
These results indicate that the key advantage of BiReT does not come merely from an additional retry, but from explicitly flipping the reasoning strategy. By switching reasoning modes, BiReT forces the model to approach the query from an entirely fresh perspective, and it is this flip, rather than resampling, that produces the correct recoveries.
Further details regarding BiReT can be found in the Supplementary Material (Sec.~\ref{supp:sec:biret}).

\begin{table}[t]
\centering
\caption{Transfer of the selected layers and threshold across proxy sources. Parameters determined on each source via the procedure of Sec.~\ref{sec:calibration} are fixed and directly applied to all benchmarks.}
\label{tab:layer_threshold_transfer}
\renewcommand{\arraystretch}{0.85}
\resizebox{\linewidth}{!}{
\begin{tabular}{l|cc|ccc|cc}
\toprule
\multicolumn{1}{c|}{\multirow{2}{*}{Method}}
& \multirow{2}{*}{Selected Layers}
& \multirow{2}{*}{Threshold ($\tau$)}
& \multicolumn{3}{c|}{ReasonSeg (gIoU)}
& \multirow{2}{*}{LogicVista}
& \multirow{2}{*}{MMVP} \\
\cmidrule(lr){4-6}
& & & val & test & Avg. & & \\
\midrule
Qwen2.5VL-7B~\cite{qwen25vl} (All-Non-CoT) & - & - & 59.9 & 58.0 & 59.0 & 37.81 & 58.00 \\
Qwen2.5VL-7B~\cite{qwen25vl} (All-CoT)     & - & - & 63.0 & 60.7 & 61.9 & 44.30 & 55.33 \\
\midrule
\rowcolor[HTML]{e0e0e0} DA-CoT \textit{(default)} & 23, 24, 25 & 6.21 & \textbf{68.0} & 64.6 & \textbf{66.3} & 45.64 & 59.33 \\
\rowcolor[HTML]{e0e0e0} DA-CoT \textit{(OK-VQA}~\cite{okvqa}\textit{)} & 16, 24, 25 & 6.14 & 67.2 & \textbf{65.0} & \underline{66.1} & 44.52 & \textbf{62.00} \\
\rowcolor[HTML]{e0e0e0} DA-CoT \textit{(A-OKVQA}~\cite{aokvqa}\textit{)} & 16, 23, 25 & 6.07 & 66.8 & \underline{64.7} & 65.8 & \underline{45.86} & 61.33 \\
\rowcolor[HTML]{e0e0e0} DA-CoT \textit{(GQA}~\cite{gqa}\textit{)} & 24, 25, 28 & 6.00 & \underline{67.7} & 64.5 & \underline{66.1} & \textbf{46.31} & \textbf{62.00} \\
\bottomrule
\end{tabular}
}
\vspace{-2mm}
\end{table}

\begin{table}[t]
\centering
\caption{Ablation study on the number of layers $k$ without Bidirectional Rethinking.}
\label{tab:layer_ablation}
\vspace{-1mm}
\renewcommand{\arraystretch}{0.85}
\resizebox{0.8\linewidth}{!}{
\begin{tabular}{c|*{2}{>{\centering\arraybackslash}p{1.5cm}}>{\centering\arraybackslash}p{1.5cm}*{2}{>{\centering\arraybackslash}p{1.5cm}}}
\toprule
ReasonSeg & \textit{$k=1$} & \textit{$k=2$} & \textit{$k=3$} & \textit{$k=4$} & \textit{$k=5$} \\
\midrule
val & 64.4 & 64.1 & 65.2 & 66.4 & 63.9 \\
test & 63.4 & 63.5 & 64.0 & 63.9 & 62.2 \\
\midrule
selected layers & 23 & 23, 25 & 23-25 & 23-26 & 22-26 \\
\bottomrule
\end{tabular}
}
\vspace{-1mm}
\end{table}

\begin{table}[t]
\centering
\caption{Comparison of recovery behavior and performance change on the ReasonSeg test set.}
\label{tab:biret_comparison}
\vspace{-1mm}
\resizebox{0.8\linewidth}{!}{
\begin{tabular}{l|ccc|c}
    \toprule
    \textbf{Method} & \textbf{UniReT ($t$=0.5)} & \textbf{UniReT ($t$=0.7)} & \textbf{UniReT ($t$=1.0)} & \textbf{BiReT (Ours)} \\
    \midrule
    \textbf{Valid Recovery Rate} & 35.4\% & 38.7\% & 35.5\% & \textbf{97.2\%} \\
    \textbf{Correct Recovery Rate} & 0.9\% & 2.8\% & 2.8\% & \textbf{22.2\%} \\
    \textbf{$\Delta$ gIoU} & +0.1 & +0.1 & +0.1 & \textbf{+0.6} \\
    \bottomrule
\end{tabular}
}
\vspace{-4mm}
\end{table}

\vspace{-2mm}
\section{Conclusion}
This work reveals that indiscriminate CoT prompting induces ``overthinking'', a critical factor that impairs the visual grounding of MLLMs.
To address this, we propose DA-CoT, a novel training-free framework that selectively triggers CoT by leveraging attention entropy as an intrinsic difficulty signal.
Enhanced with a Bidirectional Rethinking mechanism for improved robustness, our approach achieves strong performance on reasoning segmentation and generalizes to referring segmentation.
Furthermore, by bypassing unnecessary reasoning steps, DA-CoT substantially reduces inference latency. Its high compatibility ensures broad applicability across diverse MLLM architectures.
This work thus suggests a path toward more intelligent MLLMs that can self-reflect on their own cognitive state and, in turn, adaptively adjust their reasoning strategy.\par

\section{Acknowledgments}
\label{sec:acknowledgement}
This work was supported by the National Research Foundation (NRF) grant funded by the Korea government (MSIT) [RS-2025-00562400] and [RS-2022-NR068754]. The authors acknowledge the Urban Big Data and AI Institute of the University of Seoul supercomputing resources (http://ubai.uos.ac.kr) made available for conducting the research reported in this paper.

{
    \small
    \bibliographystyle{ieeenat_fullname}
    \bibliography{main}
}

\twocolumn[{%
\begin{center}
    {\Large \textbf{Attention Maps Know When to Think: \\
    Training-Free Difficulty-Aware Chain-of-Thought \\
    for Reasoning Segmentation}\par}
    \vspace{0.5em}
    \rule{0.8\linewidth}{0.4pt}\\[0.3em]
    {\normalsize \textit{Supplementary Material}}\\
    \rule{0.8\linewidth}{0.4pt}
\end{center}
\vspace{1.5em}
}]

\noindent This supplementary material provides comprehensive details and extended results that complement the main paper.
Section~\ref{supp:sec:method} details the structured CoT prompt design and its baseline validation, provides the complete mathematical formulation of the elbow detection procedure, analyzes the validity of our entropy formulation, and presents the complete DA-CoT determination and inference algorithm with Bidirectional Rethinking.
Section~\ref{supp:sec:exp} presents extended experimental results, including layer and threshold configurations across various MLLMs, the robustness of DA-CoT to proxy-set construction and in proxy-set-free environments, and a target-size control for the per-sample analysis of Sec.~\ref{sec:difficulty}.
Section~\ref{supp:sec:biret} analyzes the Bidirectional Rethinking (BiReT) mechanism, detailing its rule-based failure detection logic, recovery behavior, and overall computational efficiency.
Section~\ref{supp:sec:case_studies} offers extensive qualitative analyses, featuring additional CoT-beneficial and CoT-detrimental examples, a quantitative breakdown of CoT-induced failure modes, evidence of overthinking in CoT-optimized models, and an examination of DA-CoT's limitations in ``confidently wrong'' scenarios.

\hypersetup{linkbordercolor=black,linkcolor=black}

\renewcommand{\thesection}{\Alph{section}}
\setcounter{section}{0}

\section{Methodological Details}
\label{supp:sec:method}

\subsection{CoT Prompt Design and Baseline Validation}
\label{supp:sec:prompt}

\begin{figure}[t]
\centering
\includegraphics[width=1.0\linewidth]{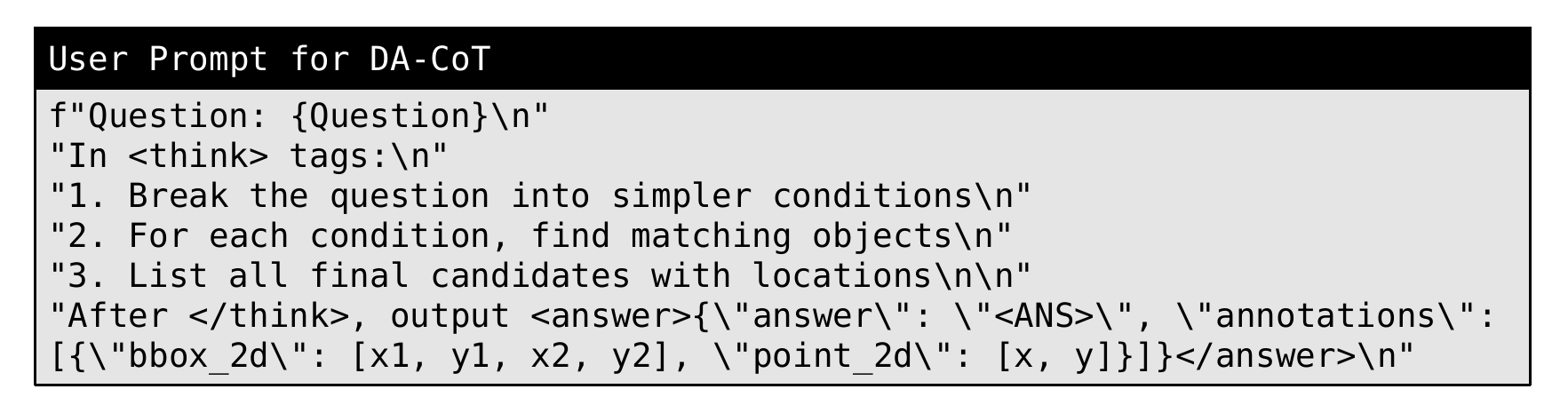}
\caption{CoT prompt for DA-CoT}
\label{fig:cot_prompt}
\end{figure}

This section provides a detailed description of the Chain-of-Thought (CoT) prompt used in our experiments and validates its effectiveness. Inspired by prior work~\cite{cot_motivation} that decomposes complex problems into smaller, manageable steps, we designed a structured prompt to systematically guide the model's reasoning process. The exact prompt used is shown in Figure~\ref{fig:cot_prompt}.

To quantitatively demonstrate the efficacy of this structured CoT prompt, we compared two all-sample settings: answering all samples without CoT prompting (All-Non-CoT) and answering all samples with our structured CoT prompt (All-CoT).

As shown in Table~\ref{tab:cot_vs_noncot_supple}, our structured CoT prompt consistently achieves higher performance than the All-Non-CoT baseline across all benchmarks. This indicates that a well-designed prompt, in itself, significantly enhances the reasoning capabilities of MLLMs and can serve as a powerful performance foundation.

Consequently, we establish the “All-CoT” result as a strong baseline for evaluating our proposed DA-CoT framework. The goal of DA-CoT, as presented in the main paper, is to surpass this strong baseline by selectively avoiding unnecessary CoT application, thereby preventing ``overthinking" and simultaneously improving both efficiency and accuracy.

\begin{table}[t]
\centering
\caption{
Quantitative validation of the structured CoT prompt design. Performance of the Qwen2.5VL-7B model is reported.
}
\vspace{-0.2cm}
\label{tab:cot_vs_noncot_supple}
\resizebox{1.0\columnwidth}{!}{
{\begin{tabular}{l|cc|ccc}
\toprule
\multicolumn{1}{c|}{\multirow{2}{*}{Method}} & \multicolumn{2}{c|}{ReasonSeg (gIoU)} & \multicolumn{3}{c}{ReferringSeg (cIoU)} \\
\cmidrule(lr){2-3} \cmidrule(lr){4-6}
 & val & test & RefCOCO & RefCOCO+ & RefCOCOg \\
\midrule
All-Non-CoT & 59.9 & 58.0 & 63.2 & 58.0 & 57.2 \\
\rowcolor[HTML]{e0e0e0} All-CoT & \textbf{63.0} & \textbf{60.7} & \textbf{73.5} & \textbf{68.2} & \textbf{60.4} \\
\bottomrule
\end{tabular}}
}
\vspace{-0.2cm}
\end{table}

\subsection{Elbow Detection Algorithm}
\label{supp:sec:elbow}

In Section~\ref{sec:method}, we compute the aggregated attention entropy for each sample by averaging the attention entropy obtained from the selected layer set $\mathcal{L}$, where the layers are chosen based on having the largest variance in attention entropy. Let $x_i$ denote the $i$-th sample in the proxy set $\mathcal{D}_{\text{proxy}}$. The aggregated attention entropy is defined as:
\begin{equation}
    \mathcal{H}_{\text{agg}}(x_i) = \frac{1}{k} \sum_{\ell \in \mathcal{L}} \mathcal{H}_{\text{attn}}^{(\ell)}(x_i),
\end{equation}
where $k = |\mathcal{L}|$ denotes the number of selected layers.

After computing the aggregated attention entropy for all samples in the proxy set $\mathcal{D}_{\text{proxy}}$, we sort them in ascending order to obtain the sorted attention entropy sequence. Let $h_i$ denote the $i$-th smallest value of $\mathcal{H}_{\text{agg}}(x)$ among all samples $x \in \mathcal{D}_{\text{proxy}}$:
\begin{equation}
    h_1 \le h_2 \le \cdots \le h_N, \quad N = |\mathcal{D}_{\text{proxy}}| .
\end{equation}

We empirically observe that the \textit{second elbow point} of the sorted attention entropy sequence serves as an effective threshold. To detect this second elbow, we first draw a straight line $L$ connecting the first point $(1, h_1)$ and the last point $(N, h_N)$. We then compute the perpendicular distance $d_i$ from each point $(i, h_i)$ to this line:
\begin{equation}
    d_i = \frac{|(N-1)(h_i - h_1) - (i-1)(h_N - h_1)|}{\sqrt{(N-1)^2 + (h_N - h_1)^2}} .
\end{equation}

The index that maximizes this perpendicular distance corresponds to the \textit{first elbow}:
\begin{equation}
    i_{\text{first}} = \underset{i \in \{1, \ldots, N\}}{\mathrm{argmax}} \, d_i .
\end{equation}

To obtain the second elbow point, we apply the same elbow-detection procedure to the tail subsequence $\{h_{i_{\text{first}}+1}, \ldots, h_N\}$. Specifically, we compute the perpendicular distance $d_j'$ from each point to the line connecting $(1, h_{i_{\text{first}}+1})$ and $(N - i_{\text{first}}, h_N)$, and determine the index that maximizes this distance:
\begin{equation}
    i_{\text{second}} = i_{\text{first}} + \underset{j \in \{1, \ldots, N - i_{\text{first}}\}}{\mathrm{argmax}} \, d_j' .
\end{equation}

Finally, the entropy threshold is set to the second elbow value:
\begin{equation}
    \tau = h_{i_{\text{second}}} .
\end{equation}

\subsection{Validity of the Entropy Formulation}
\label{supp:entropy_validity}

Our pipeline computes, for each layer $\ell$, the head-averaged attention
$\mathbf{a}^{(\ell)} = \frac{1}{H}\sum_{h=1}^{H}\mathbf{a}^{(\ell)}_h$
and its Shannon entropy $\mathcal{H}^{(\ell)}_{\text{attn}}$, then averages
these per-layer entropy values across the selected layers $\mathcal{L}$ as
$\mathcal{H}_{\text{agg}} = \frac{1}{k}\sum_{\ell \in \mathcal{L}}
\mathcal{H}^{(\ell)}_{\text{attn}}$.
Each per-head attention vector $\mathbf{a}^{(\ell)}_h$, obtained by softmax,
satisfies $\sum_{i=1}^{N_{\text{vis}}}\mathbf{a}^{(\ell)}_h[i] = 1$ by
construction. Since $\mathbf{a}^{(\ell)}$ is a convex combination of these
valid probability distributions, it also satisfies
$\sum_{i=1}^{N_{\text{vis}}}\mathbf{a}^{(\ell)}[i] = 1$, ensuring that
$\mathcal{H}^{(\ell)}_{\text{attn}}$ is well-defined. Crucially,
$\mathcal{H}_{\text{agg}}$ averages entropy \emph{scalars} across layers
rather than probability distributions, so no cross-layer distribution
mixing occurs.

We also consider an alternative \emph{per-head} formulation that computes entropy before aggregating across attention heads:
\begin{equation}
\begin{aligned}
\mathcal{H}^{(\ell)}_{\text{per-head}}
&= \frac{1}{H} \sum_{h=1}^{H} \mathcal{H}^{(\ell)}_{\text{attn},h}, \\
\mathcal{H}^{(\ell)}_{\text{attn},h}
&= -\sum_{i=1}^{N_{\text{vis}}}
\mathbf{a}^{(\ell)}_h[i]
\log \mathbf{a}^{(\ell)}_h[i].
\end{aligned}
\end{equation}
Here, Shannon entropy is computed separately for each per-head softmax distribution, and the resulting entropy values are then averaged across heads. As shown in Table~\ref{supp:tab_entropy}, this formulation selects the same layers and yields a similar threshold and comparable performance to our default formulation, indicating that the entropy signal is robust to the choice of aggregation scheme.

\subsection{DA-CoT Algorithm}
\label{supp:dacot_algorithm}
Algorithm~\ref{alg:dacot} summarizes the complete
determination and inference procedure of DA-CoT. Determination is
performed once for each MLLM backbone using only unlabeled
image--query pairs. The resulting layer set $\mathcal{L}$ and
entropy threshold $\tau$ are then fixed and reused across all
downstream datasets.
In our primary configuration, the MLLM $\mathcal{M}$ is
Qwen2.5-VL-7B, the mask generator $\mathcal{S}$ is SAM~2.1
with the Hiera-Large backbone, and the proxy set
$\mathcal{D}_{\mathrm{proxy}}$ contains 1,000 unlabeled
image--query pairs.
$\mathcal{P}_{\mathrm{CoT}}(\cdot)$ and
$\mathcal{P}_{\mathrm{NonCoT}}(\cdot)$ are given in
Sec.~A.1.

\begin{algorithm}[t]
\caption{DA-CoT Determination and Inference with Bidirectional Rethinking}
\label{alg:dacot}
\begin{algorithmic}[1]
\Require MLLM $\mathcal{M}$;
         segmentor $\mathcal{S}$;
         unlabeled proxy set $\mathcal{D}_{\mathrm{proxy}}$;
         $k=3$; CoT and Non-CoT prompts
         $\mathcal{P}_{\mathrm{CoT}}$ and
         $\mathcal{P}_{\mathrm{NonCoT}}$
\Ensure Predicted segmentation mask $\hat{M}$ or
        \textsc{Failure}

\Statex \textbf{One-time determination}
\ForAll{$x\in\mathcal{D}_{\mathrm{proxy}}$}
    \ForAll{transformer layers $\ell$}
        \State $\mathcal{H}_{\mathrm{attn}}^{(\ell)}(x)
        \gets
        \Call{AttentionEntropy}{\mathcal{M},x,\ell}$
    \EndFor
\EndFor

\State $\mathcal{L}\gets
\operatorname{TopK}_{\ell}\!\left(
\operatorname{Var}_{x\in\mathcal{D}_{\mathrm{proxy}}}
[\mathcal{H}_{\mathrm{attn}}^{(\ell)}(x)],k\right)$

\ForAll{$x\in\mathcal{D}_{\mathrm{proxy}}$}
    \State $\mathcal{H}_{\mathrm{agg}}(x)\gets
    \frac{1}{k}\sum_{\ell\in\mathcal{L}}
    \mathcal{H}_{\mathrm{attn}}^{(\ell)}(x)$
\EndFor

\State $\tau\gets
\Call{SecondElbow}{
\operatorname{sort}_{\uparrow}
\{\mathcal{H}_{\mathrm{agg}}(x):
x\in\mathcal{D}_{\mathrm{proxy}}\}}$
\Statex

\Statex \textbf{Per-sample inference}
\State Receive image--query pair $x^\star=(I^\star,q^\star)$
\State $\mathcal{H}_{\mathrm{agg}}(x^\star)\gets
\frac{1}{k}\sum_{\ell\in\mathcal{L}}
\Call{AttentionEntropy}{\mathcal{M},x^\star,\ell}$

\If{$\mathcal{H}_{\mathrm{agg}}(x^\star)\geq\tau$}
    \State $r\gets\mathrm{CoT}$
\Else
    \State $r\gets\mathrm{NonCoT}$
\EndIf

\State $y\gets
\Call{Generate}{
\mathcal{M},I^\star,\mathcal{P}_{r}(q^\star)}$

\If{\Call{Invalid}{$y$}}
    \Comment{BiReT}
    \State $r\gets\Call{Opposite}{r}$
    \State $y\gets
    \Call{Generate}{
    \mathcal{M},I^\star,\mathcal{P}_{r}(q^\star)}$
\EndIf

\If{\Call{Invalid}{y}}
    \State \Return \textsc{Failure}
\EndIf

\State $(\hat{a},\mathcal{B},\mathcal{P})
\gets\Call{ParseResponse}{y}$
\State $\hat{M}\gets
\mathcal{S}(I^\star;\mathcal{B},\mathcal{P})$
\State \Return $\hat{M}$
\end{algorithmic}
\end{algorithm}

\subsection{Mathematical Formulation of DA-CoT}
\label{supp:math_formulation}

This section provides the complete mathematical formulation of the attention-entropy signal used in DA-CoT, including attention entropy computation, layer selection, entropy aggregation, and threshold determination.

\paragraph{Attention Entropy.}
For a transformer layer $\ell$, let
$\mathbf{Q}^{(\ell)}_{\mathrm{sos}} \in \mathbb{R}^{1 \times d}$
denote the query vector corresponding to the \texttt{<SOS>} token, and let
$\mathbf{K}^{(\ell)}_{\mathrm{vis}} \in \mathbb{R}^{N_{\mathrm{vis}} \times d}$
denote the key matrix of the $N_{\mathrm{vis}}$ visual tokens.
For each attention head $h \in \{1,\ldots,H\}$, we compute the attention distribution from the \texttt{<SOS>} token to the visual tokens as
\begin{equation}
\mathbf{a}^{(\ell)}_h
=
\operatorname{softmax}
\left(
\frac{
\mathbf{Q}^{(\ell),h}_{\mathrm{sos}}
\left(\mathbf{K}^{(\ell),h}_{\mathrm{vis}}\right)^{\top}
}{
\sqrt{d/H}
}
\right),
\label{eq:attn_per_head}
\end{equation}
where $\mathbf{a}^{(\ell)}_h \in \mathbb{R}^{N_{\mathrm{vis}}}$.
We average the attention distributions across heads,
\begin{equation}
\mathbf{a}^{(\ell)}
=
\frac{1}{H}
\sum_{h=1}^{H}
\mathbf{a}^{(\ell)}_h,
\label{eq:attn_head_avg}
\end{equation}
and define the attention entropy of layer $\ell$ using Shannon entropy:
\begin{equation}
\mathcal{H}^{(\ell)}_{\mathrm{attn}}(x)
=
-\sum_{i=1}^{N_{\mathrm{vis}}}
\mathbf{a}^{(\ell)}[i]
\log \mathbf{a}^{(\ell)}[i].
\label{eq:attn_entropy}
\end{equation}
Low entropy indicates concentrated visual attention, whereas high entropy reflects a more dispersed attention distribution.

\paragraph{Entropy-Sensitive Layer Selection.}
Given an unlabeled proxy set $\mathcal{D}_{\mathrm{proxy}}$, we identify the $k$ layers whose attention entropy varies most across samples:
\begin{equation}
\mathcal{L}
=
\operatorname{TopK}_{\ell}
\left(
\operatorname{Var}_{x \in \mathcal{D}_{\mathrm{proxy}}}
\left[
\mathcal{H}^{(\ell)}_{\mathrm{attn}}(x)
\right],
k
\right).
\label{eq:layer_selection}
\end{equation}
Layers with large entropy variance are more sensitive to changes between confident and uncertain grounding behavior. In our experiments, we use $k=3$.

\paragraph{Layer-Aggregated Entropy.}
For each sample $x$, the entropy values from the selected layers are averaged to obtain a single difficulty score:
\begin{equation}
\mathcal{H}_{\mathrm{agg}}(x)
=
\frac{1}{k}
\sum_{\ell \in \mathcal{L}}
\mathcal{H}^{(\ell)}_{\mathrm{attn}}(x).
\label{eq:aggregated_entropy}
\end{equation}
This scalar is subsequently used to determine whether CoT reasoning should be activated.

\paragraph{Second-Elbow Threshold.}
We compute $\mathcal{H}_{\mathrm{agg}}(x)$ for all samples in
$\mathcal{D}_{\mathrm{proxy}}$ and sort the resulting values in ascending order:
\begin{equation}
h_1 \le h_2 \le \cdots \le h_N,
\qquad
N = |\mathcal{D}_{\mathrm{proxy}}|.
\label{eq:sorted_entropy}
\end{equation}
To identify the first elbow, we measure the perpendicular distance of each point $(i,h_i)$ from the line connecting $(1,h_1)$ and $(N,h_N)$:
\begin{equation}
d_i
=
\frac{
\left|
(N-1)(h_i-h_1)
-
(i-1)(h_N-h_1)
\right|
}{
\sqrt{(N-1)^2 + (h_N-h_1)^2}
}.
\label{eq:first_elbow_distance}
\end{equation}
The first-elbow index is then
\begin{equation}
i_{\mathrm{first}}
=
\arg\max_{i \in \{1,\ldots,N\}}
d_i.
\label{eq:first_elbow}
\end{equation}

We next apply the same procedure to the tail sequence after the first elbow.
Let
\begin{equation}
g_j = h_{i_{\mathrm{first}}+j},
\qquad
j \in \{1,\ldots,M\},
\qquad
M = N-i_{\mathrm{first}}.
\label{eq:tail_sequence}
\end{equation}
For each point $(j,g_j)$ in this subsequence, its distance from the line connecting $(1,g_1)$ and $(M,g_M)$ is
\begin{equation}
d'_j
=
\frac{
\left|
(M-1)(g_j-g_1)
-
(j-1)(g_M-g_1)
\right|
}{
\sqrt{(M-1)^2 + (g_M-g_1)^2}
}.
\label{eq:second_elbow_distance}
\end{equation}
The second-elbow index in the original sequence is
\begin{equation}
i_{\mathrm{second}}
=
i_{\mathrm{first}}
+
\arg\max_{j \in \{1,\ldots,M\}}
d'_j,
\label{eq:second_elbow}
\end{equation}
and the final entropy threshold is defined as
\begin{equation}
\tau = h_{i_{\mathrm{second}}}.
\label{eq:entropy_threshold}
\end{equation}
During inference, DA-CoT activates CoT when
$\mathcal{H}_{\mathrm{agg}}(x) \ge \tau$ and otherwise follows the direct Non-CoT path.

\paragraph{Validity of the Entropy Formulation.}
Each per-head attention vector $\mathbf{a}^{(\ell)}_h$ is obtained by softmax and therefore satisfies
\begin{equation}
\sum_{i=1}^{N_{\mathrm{vis}}}
\mathbf{a}^{(\ell)}_h[i]
=
1.
\end{equation}
Since $\mathbf{a}^{(\ell)}$ in Eq.~\ref{eq:attn_head_avg} is a convex combination of valid per-head probability distributions, it also satisfies
\begin{equation}
\sum_{i=1}^{N_{\mathrm{vis}}}
\mathbf{a}^{(\ell)}[i]
=
1,
\end{equation}
ensuring that the Shannon entropy in Eq.~\ref{eq:attn_entropy} is well-defined.
Importantly, Eq.~\ref{eq:aggregated_entropy} averages entropy \emph{scalars} across layers rather than attention distributions, so no cross-layer probability-distribution mixing occurs.

\paragraph{Alternative Per-Head Entropy.}
We additionally consider an alternative formulation that computes entropy before aggregating across attention heads:
\begin{equation}
\begin{aligned}
\mathcal{H}^{(\ell)}_{\mathrm{per\text{-}head}}
&=
\frac{1}{H}
\sum_{h=1}^{H}
\mathcal{H}^{(\ell)}_{\mathrm{attn},h}, \\
\mathcal{H}^{(\ell)}_{\mathrm{attn},h}
&=
-\sum_{i=1}^{N_{\mathrm{vis}}}
\mathbf{a}^{(\ell)}_h[i]
\log \mathbf{a}^{(\ell)}_h[i].
\end{aligned}
\label{eq:per_head_entropy}
\end{equation}
Here, Shannon entropy is computed separately for each per-head attention distribution, and the resulting entropy values are then averaged across heads.
As shown in Table~\ref{supp:tab_entropy}, this formulation selects the same layers and yields a similar threshold and comparable performance to our default formulation, indicating that the entropy signal is robust to the choice of aggregation scheme.

\begin{table}[t]
\centering
\caption{Comparison between the default head-averaged entropy and the alternative per-head entropy formulation on ReasonSeg, reported without Bidirectional Rethinking. The two formulations yield identical layer selection and comparable performance.}
\label{supp:tab_entropy}
\resizebox{\columnwidth}{!}{%
\begin{tabular}{lccc}
\toprule
\multirow{2}{*}{Formulation} & \multirow{2}{*}{Layers} & \multirow{2}{*}{$\tau$} & ReasonSeg (gIoU) \\
\cmidrule(lr){4-4}
 & & & val~/~test \\
\midrule
Default (head-averaged) & 23, 24, 25 & 6.21 & 65.2~/~64.0 \\
Per-head entropy & 23, 24, 25 & 6.16 & 65.5~/~64.1 \\
\bottomrule
\end{tabular}%
}
\end{table}

\section{Additional Experiment Results}
\label{supp:sec:exp}
\subsection{Layer and Threshold Determination}

\begin{table}[h]
\centering
\caption{Layers and thresholds selected during the layer and threshold determination phase using the proxy set.}
\label{supp:tab:precalibration_results}
\begin{tabular}{l|c|c}
\toprule

Model                & Selected Layers   & Thresholds\\
\midrule
Seg-Zero-7B~\cite{segzero}          & 8, 23, 24        & 5.95          \\
VisionReasoner-7B~\cite{visionreasoner}    & 23, 24, 25       & 6.32           \\
Qwen2.5VL-7B~\cite{qwen25vl}         & 23, 24, 25       & 6.21          \\
Qwen2.5VL-32B~\cite{qwen25vl}        & 44, 45, 58       & 5.15           \\
Qwen2.5VL-72B~\cite{qwen25vl}        & 63, 73, 74       & 5.93           \\
Qwen3VL-8B~\cite{qwen3}           & 8, 31, 33        & 5.09           \\
GLM4.1V-9B~\cite{GLM41}           & 7, 23, 26        & 5.71           \\

\bottomrule
\end{tabular}
\vspace{-1mm}
\end{table}

Table~\ref{supp:tab:precalibration_results} presents the determined layers and thresholds for various MLLMs evaluated in Table~\ref{tab:plugandplay_scalability}. These parameters are obtained using $k=3$ on the proxy set ($\mathcal{D}_{\text{proxy}}$) as described in Section~\ref{sec:calibration} of the main paper.
The selected layers exhibit different patterns depending on the model architecture. In the case of the Qwen2.5-VL series, we observe a tendency for the selected layers to shift toward later positions as the model size increases.

Furthermore, this selection naturally aligns with recent studies~\cite{kang, mllm_grounding_layer2} that highlight the mid-to-late layers of MLLMs as the critical locus for visual grounding and semantic reasoning~\cite{how}. Because these layers are responsible for specifying visual evidence—distinct from early feature processing or final decoding—high attention entropy in this specific interval serves as an intuitive and theoretically supported proxy for the model's uncertainty.

\subsection{Robustness of Proxy Set Construction}
\label{supp:proxy_robustness}

DA-CoT determines the entropy-sensitive layer set $\mathcal{L}$ and the threshold $\tau$ from a proxy set $\mathcal{D}_{\text{proxy}}$ of $1{,}000$ image--query pairs randomly sampled from segmentation training splits~\cite{refcoco, res} (Sec. \ref{sec:calibration}). To further validate the reliability of this proxy-based determination, we conduct a comprehensive robustness study across three different proxy constructions, all reported \emph{without} Bidirectional Rethinking. A desirable property of such a determination is that the resulting parameters reflect an intrinsic property of the model rather than the specific composition of the proxy. Throughout these tests, the determination procedure remains label- and mask-free: category metadata is used only to construct the proxy variants below and never enters the attention-entropy computation.

\noindent\textbf{(i) Seed variation.} We resample the proxy with five random seeds and report the mean and standard deviation of the resulting parameters and performance.

\noindent\textbf{(ii) Category-balanced proxy.} We sample uniformly across COCO categories so that all categories are equally represented in the proxy.

\noindent\textbf{(iii) Target-category-disjoint proxy.} To evaluate generalization to categories not represented in the proxy, we map ReasonSeg target descriptions to COCO categories and remove every RefCOCO/+/g pair whose target category falls in this set, yielding a $1{,}000$-sample proxy with \emph{zero} target-category overlap with the mapped evaluation categories.

As shown in Table~\ref{supp:tab_proxy}, all three settings select the identical layer set $\{23, 24, 25\}$ and a near-identical threshold, and yield gIoU comparable to the default proxy---with the category-balanced variant even slightly higher. In particular, the target-category-disjoint proxy~(iii), which shares no target category with the evaluation set, performs on par with the default. These results show that the proxy estimates a \emph{model-level} attention-dispersion signal that is independent of category composition, confirming that DA-CoT's parameter determination remains reliable across diverse proxy constructions.

\begin{table}[t]
\centering
\caption{Robustness of DA-CoT to proxy set construction on the ReasonSeg benchmark, reported without Bidirectional Rethinking. Despite varying the sampling seed, category balance, and target-category coverage, the selected layers, threshold $\tau$, and gIoU remain stable.}
\label{supp:tab_proxy}
\resizebox{\columnwidth}{!}{%
\begin{tabular}{lcccc}
\toprule
\multirow{2}{*}{Setting} & \multirow{2}{*}{Layers} & \multirow{2}{*}{$\tau$} & \multicolumn{2}{c}{ReasonSeg (gIoU)} \\
\cmidrule(lr){4-5}
 & & & val & test \\
\midrule
Default proxy & 23, 24, 25 & 6.21 & 65.2 & 64.0 \\
5-seed random & 23, 24, 25 & 6.23$\pm$0.1 & 65.0$\pm$0.3 & 64.2$\pm$0.1 \\
Category-balanced proxy & 23, 24, 25 & 6.38 & 66.3 & 64.6 \\
Target-cat.-disjoint proxy & 23, 24, 25 & 6.06 & 65.0 & 64.0 \\
\bottomrule
\end{tabular}%
}
\end{table}

\subsection{Robustness in Proxy-Set-Free Environments.}
\label{supp:proxy_free}
To investigate whether our framework remains effective even when a predefined proxy set ($\mathcal{D}_{\text{proxy}}$) is unavailable, we evaluate its performance in a completely proxy-set-free setting. This analysis is not used for our main comparisons, which use the fixed proxy set described in Sec.~\ref{sec:calibration}; it is included only to evaluate robustness when no external proxy set is available. Specifically, we dynamically determine the entropy-sensitive layers $L$ and the threshold $\tau$ by randomly sampling a mere $5\%$ of the target test set---\textbf{using solely the raw input images and text queries without any ground-truth labels}. As shown in Table~\ref{tab:dynamic}, despite relying on such a minimal and entirely unlabeled subset, this proxy-set-free approach achieves highly competitive performance compared to our default setting that utilizes the full $\mathcal{D}_{\text{proxy}}$. Furthermore, the identified layers and threshold values closely mirror the trends observed in our primary experiments. This suggests that DA-CoT can determine its layers and threshold dynamically from a small unlabeled subset when no predefined proxy set is available.

\begin{table}[t]
\centering
\caption{Comparison between DA-CoT and DA-CoT (w/o proxy set) on ReasonSeg benchmark. Performance is reported as gIoU.}
\label{tab:dynamic}
\resizebox{1.0\columnwidth}{!}{
{\begin{tabular}{l|cc|c|c}
\toprule
\multicolumn{1}{c|}{\multirow{2}{*}{Method}} & \multicolumn{2}{c|}{ReasonSeg} & \multirow{2}{*}{\shortstack{Selected \\ Layers ($L$)}} & \multirow{2}{*}{Threshold ($\tau$)} \\
\cmidrule(lr){2-3}
\multicolumn{1}{c|}{} & \multicolumn{1}{c}{val} & \multicolumn{1}{c|}{test} & & \\
\midrule
DA-CoT & 68.0 & 64.6 & 23, 24, 25 & 6.21 \\
DA-CoT (w/o proxy set) & 67.6 & 64.2 & 23, 24, 25 & 6.24 \\
\bottomrule
\end{tabular}}
}
\end{table}

\subsection{Per-Sample CoT Benefit: Target-Size Control}
\label{supp:per_sample}

\begin{figure}[t]
\centering
\includegraphics[width=\linewidth]{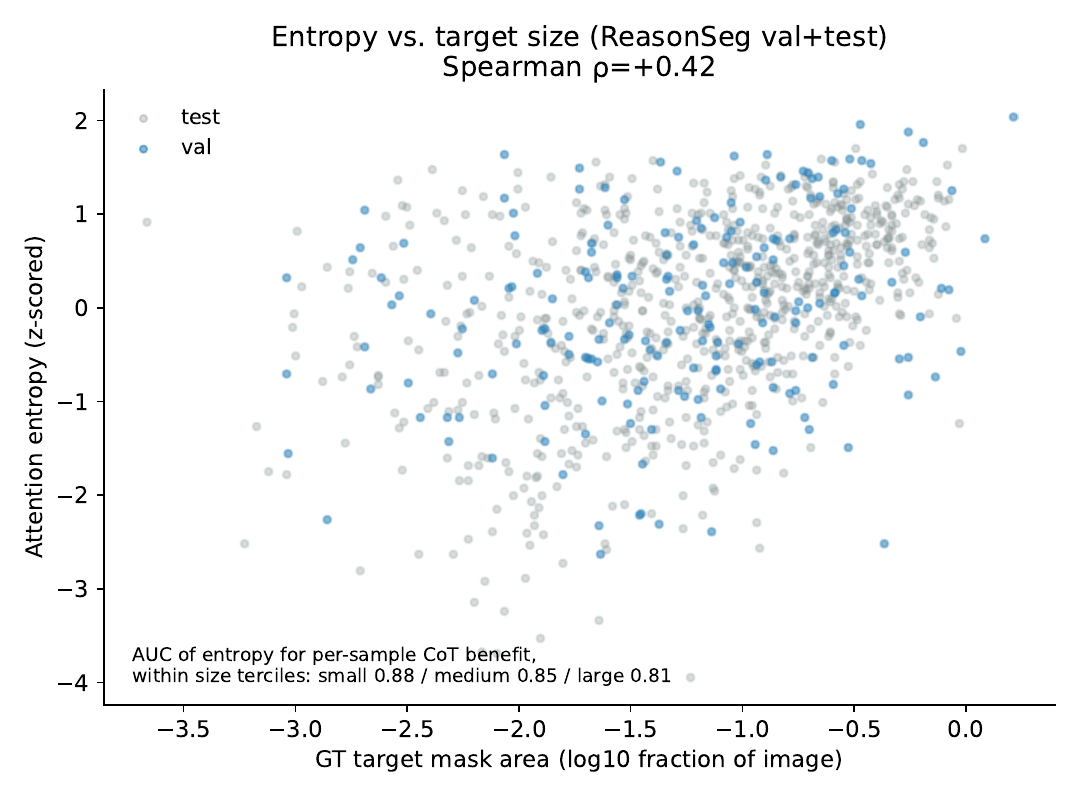}
\caption{Attention entropy vs.\ ground-truth target size on held-out ReasonSeg val+test. Annotations report the within-size-tercile AUC of entropy for per-sample CoT benefit.}
\label{fig:size_control}
\end{figure}

Since attention entropy could reflect target size rather than difficulty, we examine the two on the same held-out samples as Fig.~\ref{fig:mean_entropy}(b--d).
Entropy correlates moderately with target size (GT mask area as log fraction of the image; Spearman $\rho = +0.42$; Fig.~\ref{fig:size_control}), yet the entropy-benefit relation survives size control: the rank-based partial correlation of entropy with $\Delta\text{IoU}$ given log target size is $+0.57$, and within each size tercile entropy still predicts per-sample CoT benefit (AUC $= 0.88/0.85/0.81$ for small/medium/large targets).
As inputs are resized to $840 \times 840$, the visual-token count is constant ($N_{\text{vis}} = 900$), so normalized entropy $\mathcal{H}/\log N_{\text{vis}}$ is a fixed rescaling of $\mathcal{H}$ and leaves all rank statistics unchanged.

\subsection{Additional comparison on ReasonSeg.}
For completeness, we summarize zero-shot performance on the ReasonSeg test set in terms of both \(gIoU\) and \(cIoU\) in Table~\ref{tab:reasonseg_test}, covering a range of training-based and training-free baselines.

\begin{table}[t]
\centering
\caption{Performance comparison with zero-shot models on ReasonSeg test set. Performance is reported as gIoU and cIoU.
{$^{\dagger}$ denotes performance re-evaluated by~\cite{samr1}}.}
\label{tab:reasonseg_test}
\vspace{-0.7mm}
\resizebox{1.0\linewidth}{!}{
{\begin{tabular}{l|c|c|cc|c}
\toprule
\multicolumn{1}{c|}{Method} & Venue & Base model & gIoU & cIoU & Training \\
\midrule
LISA~\cite{lisa} & CVPR`2024 & LLaVA1.5-7B & 48.7 & 48.8 & \checkmark \\
LISA~\cite{lisa} & CVPR`2024 & LLaVA1.5-13B & 53.8 & 50.8 & \checkmark \\
Seg-Zero~\cite{segzero} & - & Qwen2.5-VL 3B & 56.1 & 48.6 & \checkmark \\
RSVP~\cite{rsvp} & ACL`2025 & Qwen2-VL 7B & 56.6 & 51.6 & $\times$ \\
RSVP~\cite{rsvp} & ACL`2025 & Gemini-Flash & 57.1 & 59.2 & $\times$ \\
Seg-Zero$^{\dagger}$~\cite{segzero} & - & Qwen2.5-VL 7B & 58.3 & 53.4 & \checkmark \\
SAM-R1~\cite{samr1} & NeurIPS`2025 & Qwen2.5-VL 7B & 60.2 & 54.3 & \checkmark \\
RSVP~\cite{rsvp} & ACL`2025 & GPT-4o & 60.3 & 60.0 & $\times$ \\
SAM-Veteran~\cite{sam-veteran} & ICLR`2026 & Qwen2.5-VL 7B & 62.6 & 56.1 & \checkmark \\
SAM3 Agent~\cite{sam3} & ICLR`2026 & Qwen2.5-VL 7B & 63.0 & 53.5 & $\times$ \\
VisionReasoner~\cite{visionreasoner} & ICLR`2026 & Qwen2.5-VL 7B & 63.6 & 55.7 & \checkmark \\
\midrule
\rowcolor[HTML]{e0e0e0} \textbf{\textbf{DA-CoT (Ours)}} & - & Qwen2.5-VL 7B & 64.6 & 58.6 & $\times$ \\
\rowcolor[HTML]{d0d0d0} \textbf{DA-CoT (Ours)} & - & Qwen2.5-VL 32B & 64.9 & 56.1 & $\times$ \\
\rowcolor[HTML]{e0e0e0} \textbf{DA-CoT (Ours)} & - & GLM4.1V-9B & 66.7 & 58.9 & $\times$ \\
\rowcolor[HTML]{d0d0d0} \textbf{DA-CoT (Ours)} & - & Qwen2.5-VL 72B & 68.0 & 60.9 & $\times$ \\
\rowcolor[HTML]{e0e0e0} \textbf{DA-CoT (Ours)} & - & Qwen3-VL 8B & 69.1 & 64.4 & $\times$ \\
\bottomrule
\end{tabular}}
}
\end{table}

\section{Bidirectional Rethinking Analysis}
\label{supp:sec:biret}

The Bidirectional Rethinking (BiReT) mechanism is designed to enhance the robustness of our framework by providing a rule-based fallback strategy. As explained in the main paper (Sec.~\ref{sec:rethinking}), BiReT is automatically activated when the initial reasoning path selected by DA-CoT fails to produce a valid output (e.g., missing coordinates or empty coordinate fields). Such failures indicate a mismatch between the estimated difficulty and the reasoning path actually required for the sample. BiReT then provides a second chance for the model to generate a valid output by issuing a \textit{flipping signal}, which switches the reasoning strategy from Non-CoT to CoT, or vice versa.

\subsection{Rule-based Failure Detection}
\label{supp:failure_detection}
\begin{figure}[t]
\centering
\includegraphics[width=\linewidth]{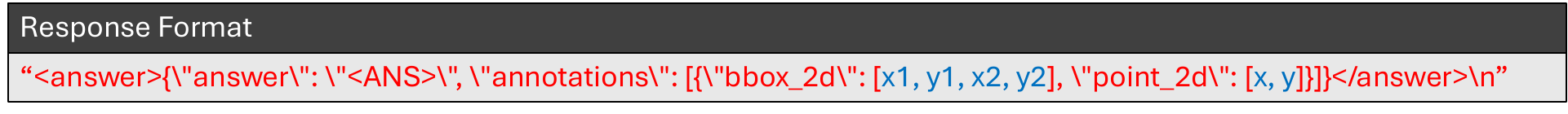}
\caption{Response format and rule-based failure detection criteria for BiReT. The mechanism triggers a flipping signal when structural components shown in \textcolor{red}{red} are incomplete or coordinate values shown in blue are missing.}
\label{fig:biret_example}
\end{figure}

To make this trigger precise, BiReT relies on a strict rule-based failure detection mechanism rather than vague criteria. The \textit{flipping signal} refers to switching the reasoning strategy from Non-CoT to CoT, or vice versa. If the flipped second pass still fails to produce a valid output, we regard the sample as a final failure of the model. As illustrated in Figure~\ref{fig:biret_example}, BiReT is triggered when:
(1) the structural components of the JSON output (marked in \textcolor{red}{red}) are incomplete, such as missing closing brackets, or
(2) the coordinate values (marked in blue) are empty, even when the JSON format itself is valid.

\begin{figure}[t]
\centering
\includegraphics[width=1.0\linewidth]{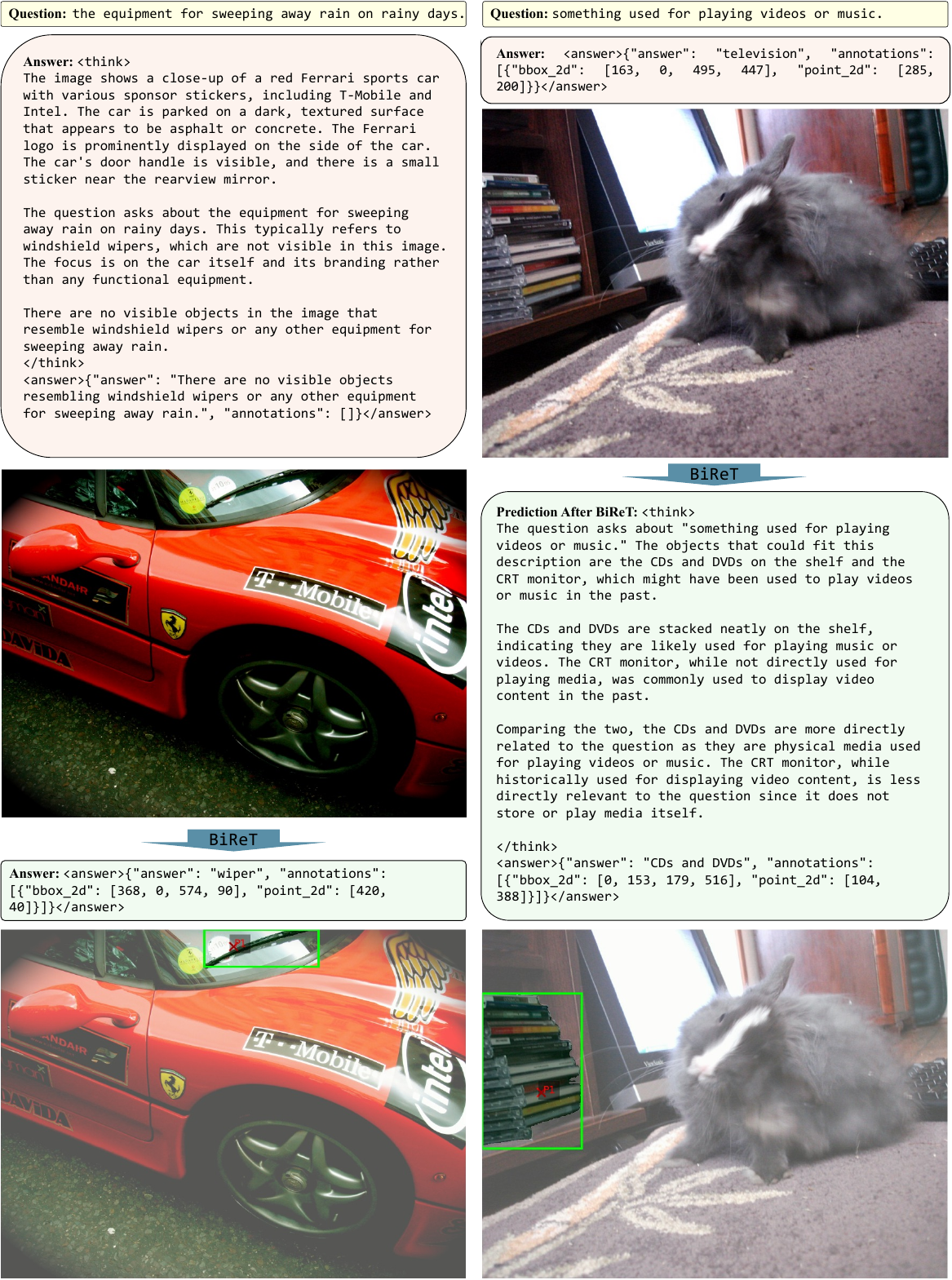}
\caption{Qualitative examples of BiReT.}
\label{fig:biret_motivation}
\end{figure}

During the design of BiReT, we empirically observed a strong correlation between instances of overthinking or underthinking and the breakdown of the output's structural integrity. As illustrated in Figure~\ref{fig:biret_motivation}, failures to complete the JSON format or omissions of bounding box coordinates serve as strong indicators that the reasoning process has derailed. This empirical finding served as the primary motivation for adopting our lightweight, rule-based failure detection logic. BiReT is executed at most once per sample.

\subsection{Activation Frequency and Recovery Behavior}

To quantify the practical effectiveness and efficiency of this mechanism, we analyzed its activation frequency and recovery behavior on the ReasonSeg benchmark.

\begin{table}[t]
\centering
\small
\caption{
Efficiency and Efficacy of the BiReT Mechanism.
Both recovery rates are computed over the triggered samples.
}
\label{tab:BiReT_supple}
\begin{tabular}{lcc}
\toprule
Metric & val & test \\
\midrule
Trigger Rate (\%) & 5.0 & 4.6 \\
Valid Recovery Rate (\%) & 80.0 & 97.2 \\
Correct Recovery Rate (\%) & 70.0 & 22.2 \\
\bottomrule
\end{tabular}
\vspace{-0.2cm}
\end{table}

\begin{table}[t]
\centering
\caption{
Performance and latency are measured on the ReasonSeg Benchmark.
}
\label{tab:biret_latency}
\resizebox{0.6\columnwidth}{!}{
{\begin{tabular}{l|cc|c}
\toprule
\multicolumn{1}{c|}{Method} & \multicolumn{2}{c|}{ReasonSeg (gIoU)} & \multirow{2}{*}{time/img (sec)} \\
\cmidrule(lr){2-3}
 & val & test & \\
\midrule
Qwen 2.5-VL 7B & 63.0 & 60.7 & 9.9 \\
\rowcolor[HTML]{e0e0e0} + DA-CoT & 65.2 \scriptsize \textcolor{red}{($\uparrow$ 2.2)} & 64.0 \scriptsize \textcolor{red}{($\uparrow$ 3.3)} & 4.4 \\
\rowcolor[HTML]{e0e0e0} + BiReT & 68.0 \scriptsize \textcolor{red}{($\uparrow$ 2.8)} & 64.6 \scriptsize \textcolor{red}{($\uparrow$ 0.6)} & 4.7 \\
\bottomrule
\end{tabular}}
}
\vspace{-0.2cm}
\end{table}

As shown in Table~\ref{tab:BiReT_supple}, BiReT is triggered for about 5\% of the total samples, exactly those where the selected reasoning mode fails to produce a valid output, and flipping the reasoning strategy restores a valid output for 97.2\% of them on test (80.0\% on val). As reported in Table~\ref{tab:biret_comparison}, the restored output is also correct (IoU $>$ 0.5) for 22.2\% of the initially failed test samples (70.0\% on val), whereas UniReT never exceeds 2.8\%: the strategy flip, not the retry itself, is what produces the correct recoveries. Since a training-free mechanism cannot add knowledge to the backbone, we view BiReT primarily as an output-reliability component that provides a usable prediction at $+0.3$s latency, contributing a further $+0.6$ test gIoU on top of DA-CoT.

\subsection{Design Motivation and Efficiency}

Admittedly, one could consider more sophisticated verification methods as the trigger condition for BiReT. For instance, the model could be prompted to perform self-correction on its generated reasoning chain and predicted regions via an additional forward pass, or a more capable external model could be deployed as a ``Judge'' to evaluate the correctness of the initial output. However, such approaches inevitably incur substantial computational costs and latency overheads.

In contrast, our lightweight, rule-based BiReT demonstrates its efficiency, restoring a valid output for up to 97.2\% of initially failed samples purely through strategy switching, without requiring any additional verification forward passes or heavy external models. Consequently, this simple rule-based mechanism strikes an optimal balance, simultaneously maximizing both the computational efficiency and output reliability of our framework. As quantitatively demonstrated in Table~\ref{tab:biret_latency}, integrating BiReT incurs a minimal latency overhead of 0.3 seconds per image (from 4.4s to 4.7s), successfully preserving the substantial speedup achieved by DA-CoT.

\section{Qualitative Analysis and Case Studies}
\label{supp:sec:case_studies}
\subsection{Further Examples of CoT-Beneficial and CoT-Detrimental Cases}
To further substantiate the core claim of our paper—that the indiscriminate use of Chain-of-Thought prompting can induce overthinking and necessitates an adaptive approach—we present additional qualitative examples where CoT is either beneficial or detrimental to the reasoning process.

\begin{figure*}[t]
\centering
\includegraphics[width=0.75\linewidth]{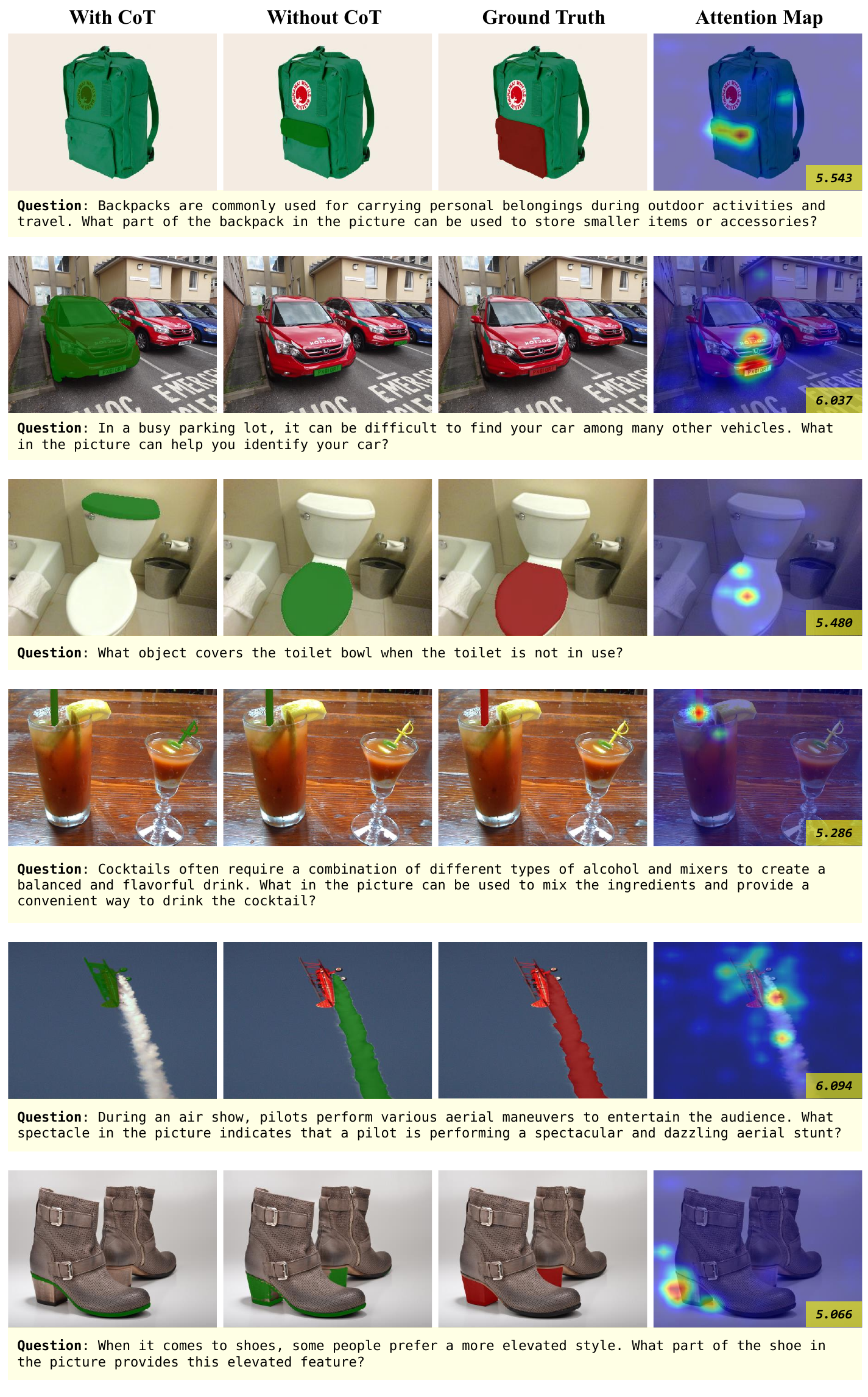}
\caption{CoT-Detrimental Cases}
\label{fig:supple_detrimental}
\end{figure*}

\noindent \textbf{CoT-Detrimental Cases}
Figure~\ref{fig:supple_detrimental} provides additional examples of CoT-detrimental cases where CoT prompting leads to performance degradation. As noted in the main paper, these are instances where the model already exhibits high confidence (indicated by low attention entropy) in localizing the correct region without CoT. However, forcibly applying CoT induces an unnecessary and convoluted reasoning process (overthinking). This can dilute the model's focus or lead it down an incorrect logical path, ultimately resulting in a misgrounded prediction.

\begin{figure*}[t]
\centering
\includegraphics[width=0.8\linewidth]{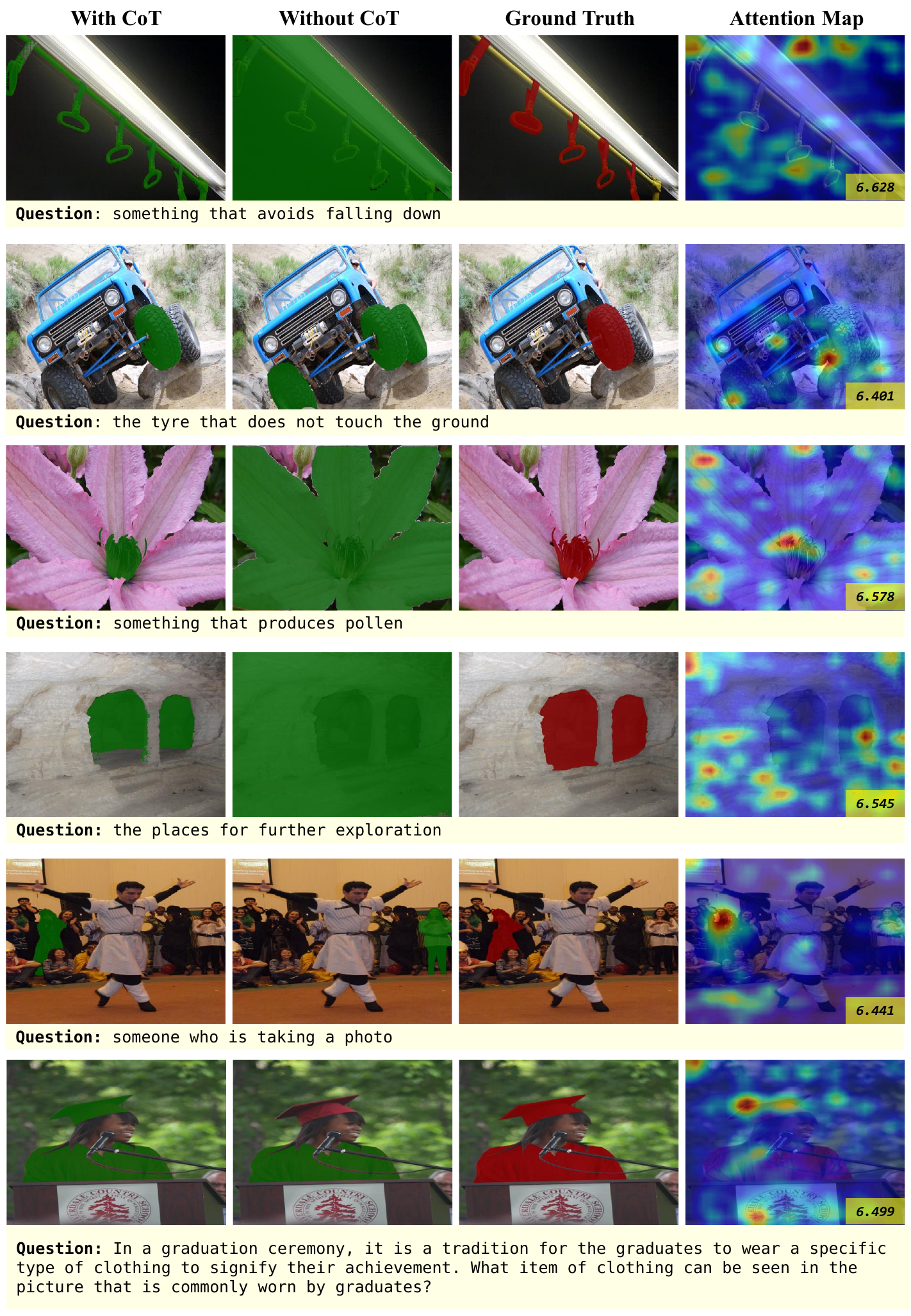}
\caption{CoT-Beneficial Cases}
\label{fig:supple_beneficial}
\end{figure*}

\noindent \textbf{CoT-Beneficial Cases}
Conversely, Figure~\ref{fig:supple_beneficial} presents additional CoT-beneficial cases where CoT plays a crucial role in improving reasoning performance. These examples typically arise when a query is challenging for the MLLM to ground or when visual cues are ambiguous, causing the model to initially exhibit high uncertainty (reflected by high attention entropy). In such scenarios, CoT guides the model to break down the problem into intermediate steps and follow a structured reasoning process. This guidance is decisive in helping the model accurately identify the correct target region.

\subsection{Quantitative Analysis of CoT-Induced Failures}
\label{supp:failure_breakdown}

Beyond the qualitative case studies of CoT-detrimental examples, we provide a quantitative error analysis that characterizes \emph{how} indiscriminate CoT prompting degrades grounding. We focus on \emph{CoT-detrimental cross-over cases}: samples that the model grounds correctly without CoT ($\mathrm{IoU}_{\mathrm{NonCoT}} > 0.5$) but fails once CoT is applied ($\mathrm{IoU}_{\mathrm{CoT}} \le 0.5$). Such cases isolate the failures directly attributable to overthinking, since the only change between the two runs is the activation of CoT. We randomly sample 165 such cross-over cases pooled from the ReasonSeg val and test splits, and assign each to its dominant failure cause: \textbf{spatial confusion} (the prediction drifts to a wrong instance during spatial reasoning), \textbf{semantic drift} (the prediction shifts to a semantically related but incorrect object), \textbf{visual hallucination} (the model grounds to an object that is not present), and \textbf{format errors / others} (malformed or incomplete output structure).

As shown in Table~\ref{supp:tab_failure}, spatial confusion is by far the dominant failure mode (50.3\%), followed by semantic drift (18.8\%), visual hallucination (16.4\%), and format errors (14.5\%). This indicates that the unnecessary reasoning steps introduced by CoT most frequently derail grounding through spatial-logic errors rather than perceptual mistakes. The result quantitatively substantiates our central claim that overthinking compromises visual grounding, and shows that this degradation manifests through structured, spatially-grounded failures specific to pixel-level grounding---directly motivating our difficulty-aware, adaptive CoT gating.

\begin{table}[t]
\centering
\caption{Failure-mode breakdown of 165 randomly sampled CoT-detrimental cross-over cases ($\mathrm{IoU}_{\mathrm{NonCoT}} > 0.5 \wedge \mathrm{IoU}_{\mathrm{CoT}} \le 0.5$) on ReasonSeg. Spatial confusion is the dominant failure mode, indicating that unnecessary reasoning most often derails grounding through spatial-logic errors.}
\label{supp:tab_failure}
\begin{tabular}{lc}
\toprule
Failure Cause & Proportion (\%) \\
\midrule
Spatial confusion & 50.3 \\
Semantic drift & 18.8 \\
Visual hallucination & 16.4 \\
Format errors / others & 14.5 \\
\bottomrule
\end{tabular}
\end{table}

\subsection{Overthinking in CoT-Optimized Models}
\begin{figure}[t]
\centering
\includegraphics[width=1.0\linewidth]{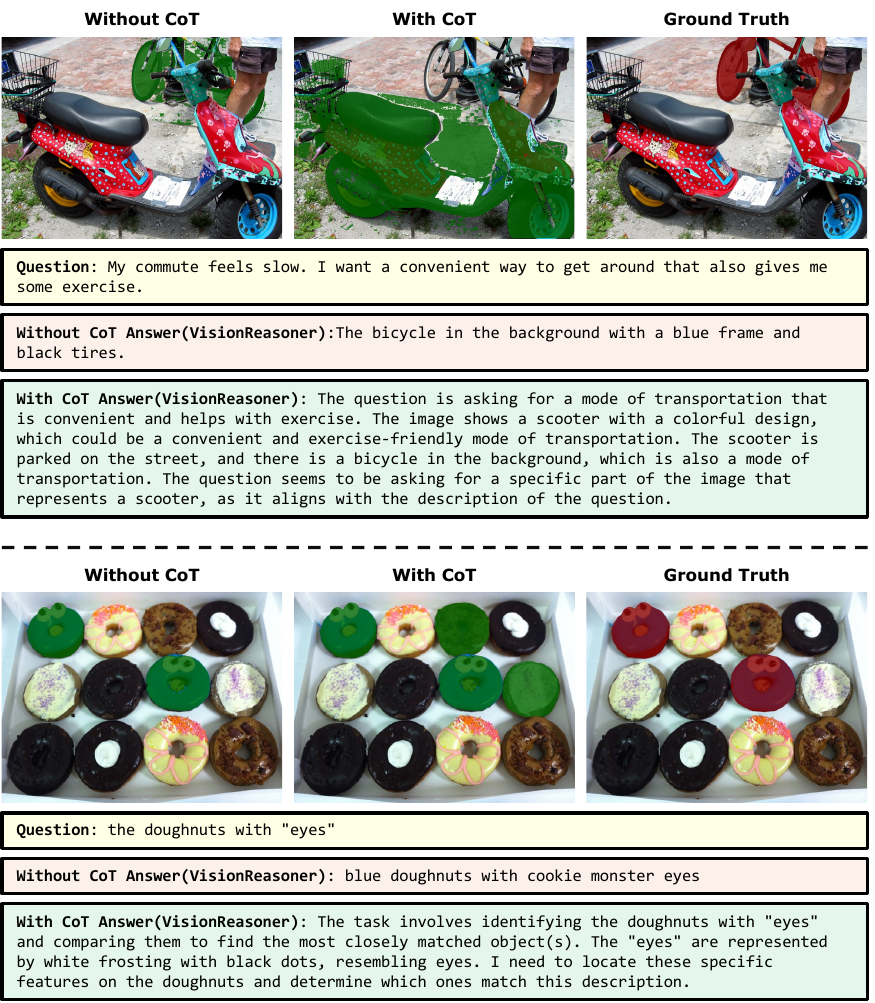}
\caption{Overthinking of VisionReasoner}
\label{fig:visionreasoner_overthinking}
\end{figure}

Even models specifically trained for CoT reasoning, such as VisionReasoner~\cite{visionreasoner}, are not immune to the problem of `overthinking.' Figure~\ref{fig:visionreasoner_overthinking} illustrates that for queries the model can answer correctly without CoT, enforcing a reasoning process can be counterproductive. As seen in the figure, the forced CoT process induces unnecessary reasoning steps that lead to model failure. This demonstrates an instance of `overthinking,' where the enforced reasoning process itself contributes to error, even though the model inherently possesses correct grounding and reasoning capabilities. By mitigating such failure modes, our method leads to performance improvements even when applied to CoT-optimized models, as is quantitatively confirmed in Table~\ref{tab:plugandplay_scalability} of the main paper. This observation highlights the potential value of an adaptive mechanism like DA-CoT, which dynamically selects the reasoning path.

\subsection{Failure Cases and Limitations of DA-CoT}
The effectiveness of DA-CoT is predicated on the model's attention entropy, and a limitation arises when this intrinsic signal does not perfectly reflect the model's true understanding. The key limitation we identified is the 'confidently wrong' scenario, where the model focuses on an incorrect region with high confidence (i.e., low entropy).

\begin{figure}[t]
\centering
\includegraphics[width=1.0\linewidth]{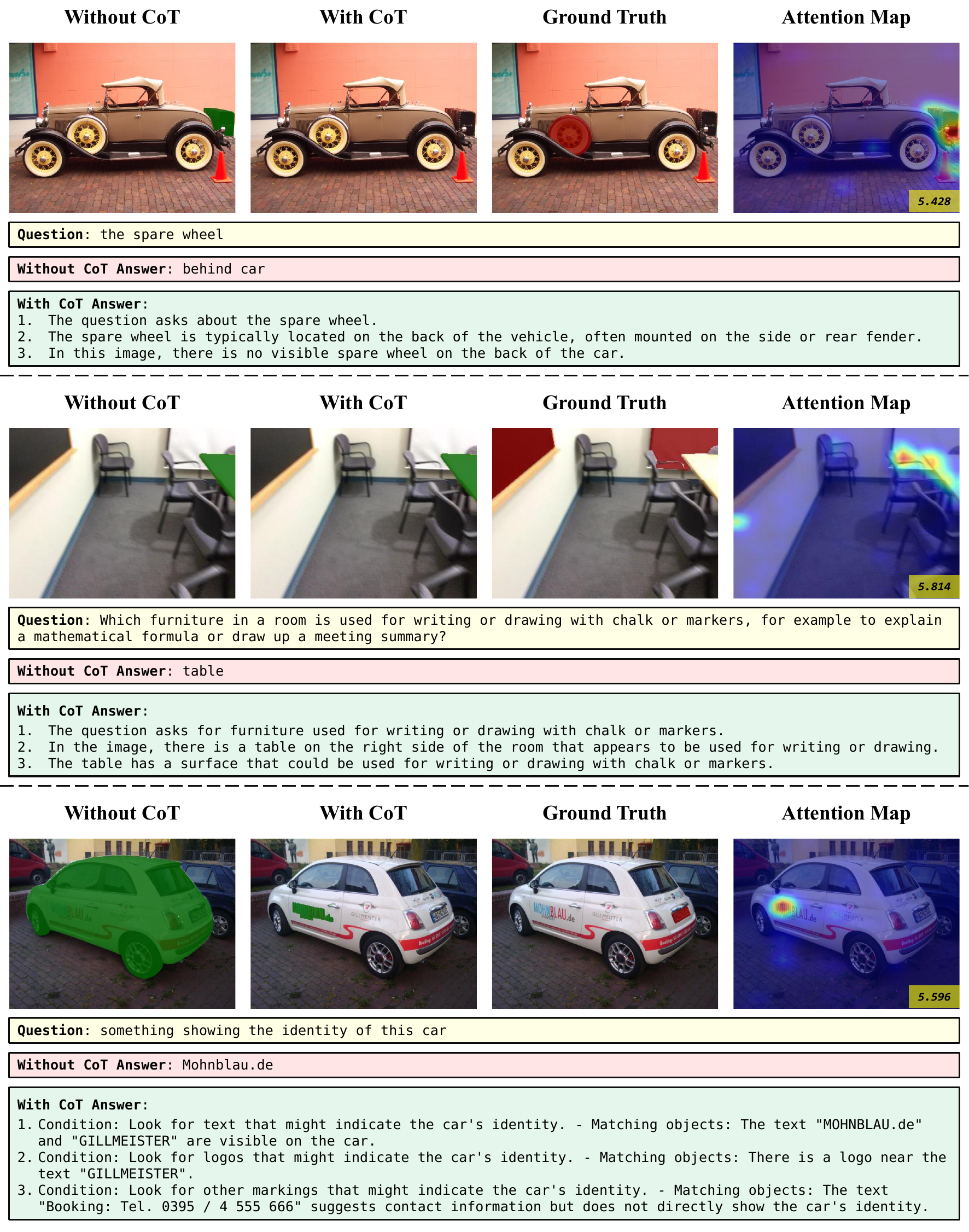}
\caption{Failure Cases of DA-CoT}
\label{fig:failure_cases}
\end{figure}

As shown in Figure~\ref{fig:failure_cases}, the model's attention map is sharply localized on a specific region, yielding a low entropy value. Based on this signal, DA-CoT judges the model to be highly confident and, as designed, does not activate Chain-of-Thought. Consequently, the model performs a direct prediction based on its flawed initial grounding, leading to an incorrect answer.

On the surface, this might appear to be an issue of incorrect reasoning path selection. However, our experiments confirmed that even when CoT was forcibly applied in such cases, the MLLM still failed to arrive at the correct answer.

This suggests that the issue is less about `reasoning path selection' and more related to a fundamental limitation in the MLLM's own perceptual capabilities. This includes cases where the model initially misinterprets the query or image, or lacks the prerequisite knowledge about the scene.

This analysis clarifies both the limitations and the scope of DA-CoT's contribution. As a training-free framework, DA-CoT cannot expand a model's fundamental perceptual abilities or directly correct its recognition errors. Instead, its core contribution lies in maximizing a model's potential performance by preventing unnecessary reasoning and selectively activating CoT only when needed.

\clearpage

\end{document}